\documentclass[sigconf,noncam]{acmart}

\AtBeginDocument{%
  }

\usepackage{multirow}
\usepackage{diagbox}
\setcopyright{none}
\renewcommand\footnotetextcopyrightpermission[1]{}
\begin{document}

\title{Distilling Vision-Language Foundation Models: A Data-Free Approach via Prompt Diversification}

\author{Yunyi Xuan}
\authornote{Equal contribution.}
\affiliation{%
  \institution{Hikvision Research Institute}
  \city{Hangzhou}
  \country{China}
}
\email{xuanyunyi@hikvision.com}
\orcid{0009-0000-6365-5475}

\author{Weijie Chen}
\authornotemark[1]
\authornote{Corresponding authors.}
\affiliation{%
  \institution{College of Computer Science and Technology, Zhejiang University \& Hikvision Research Institute}
  \city{Hangzhou}
  \country{China}
}
\email{chenweijie@zju.edu.cn}
\orcid{0000-0001-5508-473X}

\author{Shicai Yang}
\affiliation{%
  \institution{Hikvision Research Institute}
  \city{Hangzhou}
  \country{China}
}
\email{yangshicai@hikvision.com}
\orcid{0000-0002-9260-1334}

\author{Di Xie}
\authornotemark[2]
\affiliation{%
  \institution{Hikvision Research Institute}
  \city{Hangzhou}
  \country{China}}
\email{xiedi@hikvision.com}
\orcid{0000-0001-8065-5901}

\author{Luojun Lin}
\affiliation{%
  \institution{Fuzhou University}
  \city{Fuzhou}
  \country{China}}
\email{linluojun2009@126.com}
\orcid{0000-0002-1141-2487}

\author{Yueting Zhuang}
\authornotemark[2]
\affiliation{%
  \institution{College of Computer Science and Technology, Zhejiang University}
  \city{Hangzhou}
  \country{China}}
\email{yzhuang@zju.edu.cn}
\orcid{0000-0001-9017-2508}

\renewcommand{\shortauthors}{Yunyi Xuan et al.}

\begin{abstract}
Data-Free Knowledge Distillation (DFKD) has shown great potential in creating a compact student model while alleviating the dependency on real training data by synthesizing surrogate data. However, prior arts are seldom discussed under distribution shifts, which may be vulnerable in real-world applications. Recent Vision-Language Foundation Models, e.g., CLIP, have demonstrated remarkable performance in zero-shot out-of-distribution generalization, yet consuming heavy computation resources. In this paper, we discuss the extension of DFKD to Vision-Language Foundation Models without access to the billion-level image-text datasets. The objective is to customize a student model for distribution-agnostic downstream tasks with given category concepts, inheriting the out-of-distribution generalization capability from the pre-trained foundation models. In order to avoid generalization degradation, the primary challenge of this task lies in synthesizing diverse surrogate images driven by text prompts. Since not only category concepts but also style information are encoded in text prompts, we propose three novel Prompt Diversification methods to encourage image synthesis with diverse styles, namely Mix-Prompt, Random-Prompt, and Contrastive-Prompt. Experiments on out-of-distribution generalization datasets demonstrate the effectiveness of the proposed methods, with Contrastive-Prompt performing the best.
\end{abstract}


\received{20 February 2023}
\received[revised]{12 March 2023}
\received[accepted]{26 July 2023}

\maketitle

\section{Introduction}
\begin{figure}[!t]
	\centering
	\includegraphics[width=0.95\columnwidth]{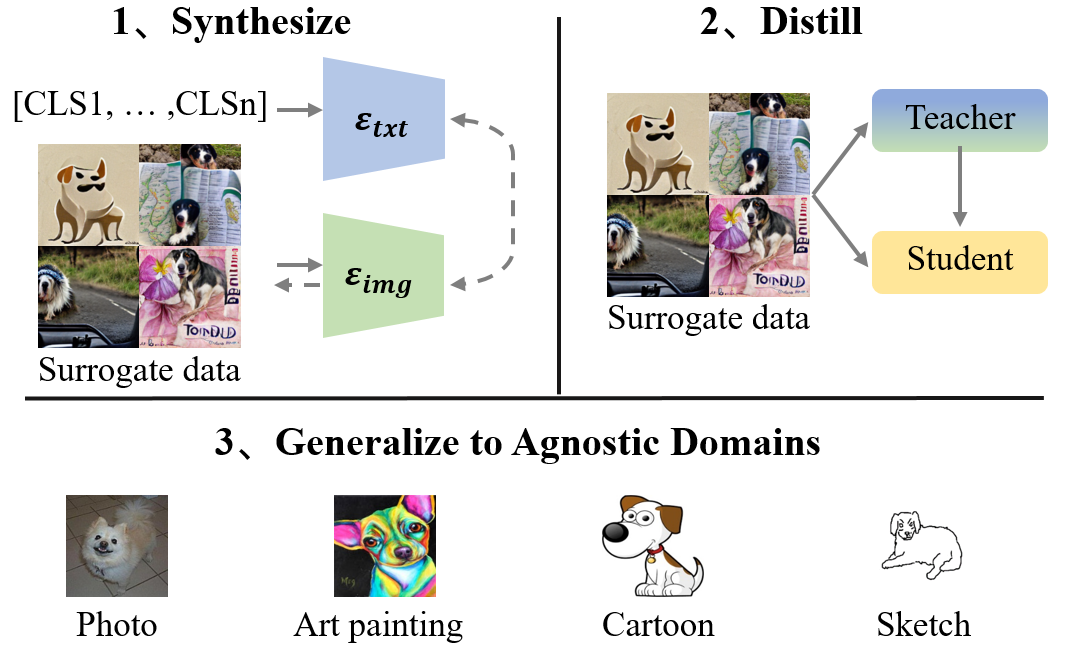} 
 	\caption{Synthesize, distill, and then generalize. Here we take CLIP as an example, including an image encoder $\mathcal{E}_{img}$ and a text encoder$\mathcal{E}_{txt}$. The surrogate images are first synthesized from the foundation model with several words of the target categories [{\tt CLS1}, ..., {\tt CLSn}] provided. Knowledge distillation is then performed upon the synthesized dataset to customize a generalizable student, wherein CLIP acts as Teacher.}
	\label{fig: application}
\end{figure}
\begin{figure*}
\centering
	\includegraphics[width=2.1\columnwidth]{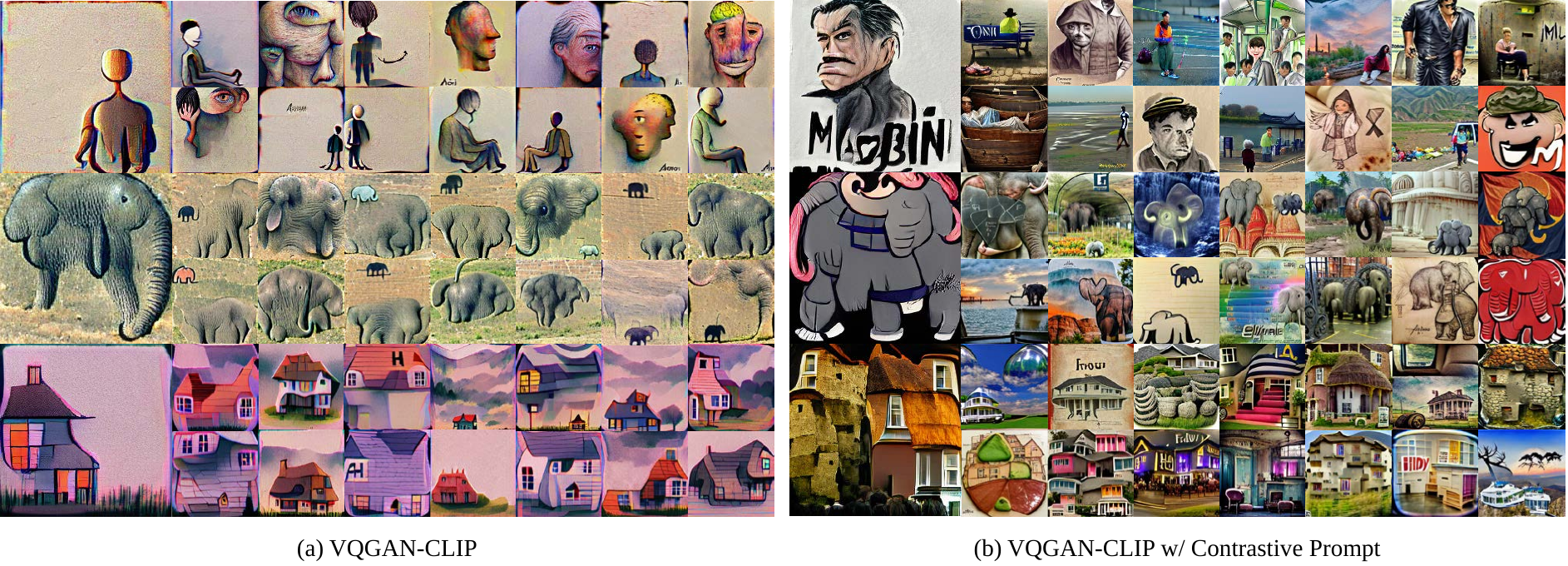} 
	\caption{A comparison of surrogate image synthesis. \emph{top:} ``person'', \emph{middle:} ``elephant'', \emph{bottom:} ``house''. Directly using VQGAN-CLIP suffers from model collapse. Contrastive-Prompt can significantly increase the data diversity with complex contexts, facilitating the process of Data-Free Knowledge Distillation from Vision-Language Foundation Models.}
	\label{vis}
\end{figure*}
\begin{figure}[!t]
	\centering
	\includegraphics[width=0.95\columnwidth]{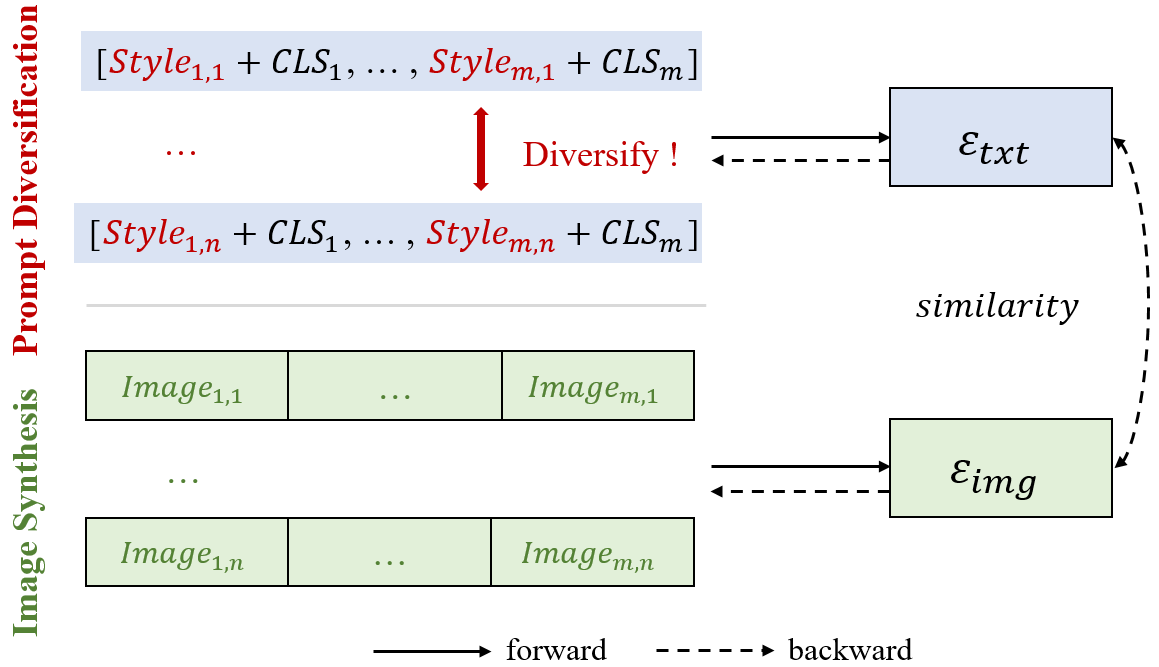} 
	\caption{Conventional DFKD aims to diversify images directly. In contrast, in the context of vision-language foundation models, we aim to diversify text prompts as a bridge to synthesize diverse surrogate images since the style information can be encoded in the text prompts implicitly. Here $m$ denotes the category number while $n$ is the sample number.}
	\label{fig: prompt-diversify}
\end{figure}
The emergence of deep neural networks has led to rapid advances in deep learning \cite{chen2019all,chen2020unsupervised}. These models can be applied to a wide range of downstream tasks, but their ever-growing model parameters inhibit applications on resource-constraint edge devices. Knowledge distillation \cite{2015KD,DIST,decoupledKD} is one of the most prevalent paradigms to solve this problem. Despite the promising progress, these approaches rely heavily on the original data, which is not always available due to copyrights and privacy concerns. To overcome this challenge, a few attempts have been made to distill knowledge without using real data, known as Data-Free Knowledge Distillation (DFKD)~\cite{Deepinv,CMI,largeDFAD,DFAD}. Specifically, DFKD adopts a two-step paradigm, including image synthesis and knowledge distillation. Assumed that prior data distribution is implicitly encoded in the teacher model. The surrogate images are first generated by inverting the pre-trained teacher model, which are then utilized in knowledge distillation as an alternative to the original data. 

Nevertheless, none of the existing works explore DFKD under distribution shifts, which are commonly encountered in real-world applications \cite{sun2022dynamic,meng2022attention}. It naturally comes to the demand for a lightweight generalizable model. From another perspective, the recent pioneer Vision-Language Foundation Models~\cite{align}, e.g., CLIP\cite{clip,huang2022transductive}, have been witnessed remarkable success in zero-shot out-of-distribution (OOD) generalization to numerous domain-agnostic downstream tasks, yet suffering from cumbersome parameters and the inaccessible original dataset. In this paper, we take a further step to explore a more ambitious version of DFKD in OOD scenarios - to customize a generalizable student from these foundation models which can be robust to any domain shift. We refer to this task as Data-Free Knowledge Distillation from Vision-Language Foundation Models (DFKD-VLFM). Unlike conventional DFKD approaches, the knowledge of publicly-available pre-trained foundation models do not need to be fully utilized. For example, CLIP \cite{clip} has the ability to recognize various category concepts, but for downstream tasks in some constrained scenarios, only a limited category space is required. Hence, as shown in Fig.\ref{fig: application}, we attempt to create a lightweight student model inheriting CLIP's generalizablity for domain-agnostic downstream tasks, based on the specific category concepts required for the tasks.

Despite several studies \cite{Deepinv,largeDFAD,FastDFKD} experimenting with DFKD on ImageNet, it remains a challenge to invert such a large-scale model to synthesize surrogate training data. Recently, the developed text-to-image models\cite{clipvqgan,DALL-E} have made significant strides in image generation, yielding high-quality images based on text prompts. In this paper, we adopt one of the most representative text-to-image models, VQGAN-CLIP, as a baseline framework to explore DFKD-VLFM. To create a generalizable student, synthetic data should cover the distribution of downstream tasks as much as possible. As shown in Fig.~\ref{fig: prompt-diversify}, in the context of vision-language foundation models, diversifying text prompts can serve as a bridge to diversify synthesized images. However, hand-crafting prompt engineering is laborious and empirical. To this end, we aim to develop a Prompt Diversification method that can generate diverse text prompts and implicitly drive the synthesis of diverse surrogate images (Fig.~\ref{vis}), leading to a generalizable student. Three attempts have been made in this paper:

\textbf{Mix-Prompt.} It is intuitive to build a dictionary of style words sourced from the Internet and randomly sample various styles from the dictionary to construct diverse text prompts. To enrich the styles, we create a series of novel styles via a random convex interpolation among different styles in the text embedding space.

\textbf{Random-Prompt.} Albeit interpolating the style dictionary, it may fail to yield instances with different postures, appearances and breeds. To create novel styles of the target category, we replace the explicit style word with a pseudo word, which is generated randomly in the textual embedding without using the style dictionary.

\textbf{Contrastive-Prompt.} 
To further ensure inter-sample diversity, we introduce contrastive learning on the instance level to optimize text prompts. Specifically, the generated text prompt is forced to differ from those in the mini-batch and the historical ones saved in the memory bank.

To thoroughly evaluate the robustness of the distilled student driven by the proposed Prompt Diversification methods, we conduct experiments on multiple domain generalization datasets, including PACS~\cite{pacs}, VLCS~\cite{vlcs}, VisDA~\cite{visda} and ImageCLEF-DA~\cite{imageclef}. Extensive experiments demonstrate a significant performance improvement against the baseline method. The contributions are summarized as follows:
\begin{itemize}
    \item We propose an interesting task in this paper termed Data-Free Knowledge Distillation from Vision-Language Foundation Models. This is an important direction to leverage the vast knowledge in the foundation models to enhance downstream tasks.
    \item To enrich the diversity of the synthesized images for knowledge distillation, we take text prompts as a bridge to describe image diversity and propose three novel Prompt Diversification methods, including Mix-Prompt, Random-Prompt, as well as Contrastive-Prompt.
    \item Extensive experiments confirm the effectiveness of the proposed methods, which can customize various task-specific models in a data-free manner for zero-shot classification or few-shot fine-tuning.
\end{itemize}

\section{Related Works}
\subsection{Vision-Language Foundation Models}
The field of AI has seen a surge of interest in vision-language pre-training, which aims to develop generic cross-modal representation to address various downstream tasks. CLIP~\cite{clip} and ALIGN~\cite{align} use image-text contrastive learning with noisy text supervision on billion-level training data to learn cross-modal representation, achieving impressive zero-shot capability. Following their footsteps, DeCLIP~\cite{Inoue2018CrossDomainWO}, SLIP~\cite{Mu2022SLIPSM} and FILIP~\cite{Yao2022FILIPFI} have further advanced this field by modifying the optimization objectives. Florence~\cite{Yuan2021FlorenceAN} trains on a vast amount of data and uses a unified model to solve extensive vision tasks. However, these models usually consist of an enormous number of parameters, limiting their practical applications. Additionally, the forbidden access to the billion-level training data precludes conventional knowledge distillation. Also, in most constrained scenarios, it is too redundant for a student to acquire all of the teacher's knowledge.

In this paper, we propose a new task called DFKD-VLFM that focuses on transferring the generalizable knowledge from the foundation models to customize lightweight models with a given task-specific label space.

\subsection{Data-Free Knowledge Distillation}
As the growing concerns about data privacy, DFKD~\cite{dafl,Deepinv,DFAD,CMI,largeDFAD,FastDFKD} has been a flourishing topic in recent years. This approach extracts knowledge from a pre-trained teacher by synthesizing images that adhere to the teacher network statistics through enforcing the BN regularization~\cite{Deepinv}. The student is then optimized by knowledge distillation upon the synthetic dataset. While DeepInversion\cite{Deepinv} generates class-conditional images from random noise, GAN~\cite{GAN}-based adversarial training has become a dominant scheme for DFKD with a generator trained from scratch~\cite{dafl,DFAD,CMI}. CMI~\cite{CMI} argues that data diversity is beneficial for distillation performance and introduces contrastive learning to increase instance discrimination. Nevertheless, the out-of-domain (OOD) problem is unstudied in the conventional DFKD, in which the student is evaluated on the dataset with the same data distribution of training data. In this paper, we address a more rigorous DFKD problem, named DFKD-VLFM, where a pioneering vision-language pre-trained model (\emph{e.g.}, CLIP~\cite{clip}) is inverted to achieve a generalizable student. Despite the promising performance of GAN-based DFKD methods, inverting a model pre-trained on a large-scale dataset remains challenging. Inspired by the remarkable advantages of recent text-to-image models~\cite{clipvqgan,DALL-E}, we take VQGAN-CLIP \cite{clipvqgan} as a baseline framework. As demonstrated that a myriad of various images can be generated by the text-to-image models with diverse text prompts, in this paper, we propose three types of prompt diversification to achieve generalizable knowledge transfer.

\section{Method}
\begin{figure*}[!t]
	\centering
	\includegraphics[width=2.1\columnwidth]{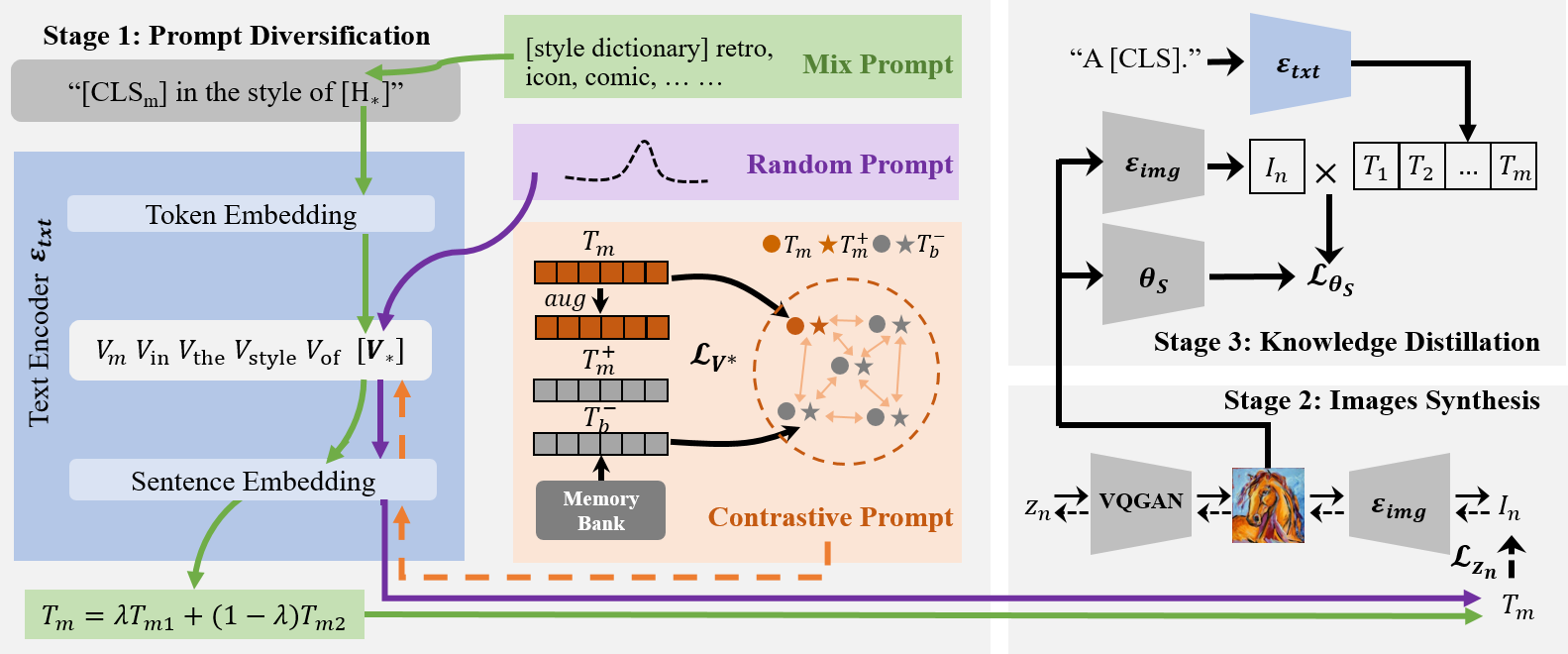} 
	\caption{Three Prompt Diversification methods, Mix-Prompt, Random-Prompt, and Contrastive-Prompt, are utilized to generate diverse text prompts $T_m$, resulting in diverse task-specific high-fidelity images. With these surrogate training images, we can customize a task-specific student model ($\theta_S$) by extracting knowledge from CLIP ($\mathcal{E}_{img}$ + $\mathcal{E}_{txt}$).}
	\label{fig: framework}
\end{figure*}
\subsection{Preliminary}
We first give the definition of DFKD-VLFM: given a pre-trained vision-language foundation model (e.g., CLIP) as the teacher $\mathcal{T}(\cdot, \theta_T)$ and a few words of the target category [{\tt CLS1}, ..., {\tt CLSn}], the objective of DFKD-VKFM is to develop a student $\mathcal{S}(\cdot, \theta_S)$ that can recognize the target categories and can be generalized to any domain-agnostic downstream task without using any real data. As an enhanced version of DFKD, the process of DFKD-VLFM can be similarly decoupled into an image synthesis step and a knowledge distillation step. Unless explicitly specified, we use CLIP as the pre-trained teacher model in this paper. During image synthesis, as mentioned above, we replace the plain generator of the conventional DFKD with the pre-trained VQGAN $\mathcal{G}(\cdot)$ to achieve high-fidelity image synthesis. Given a category-specific text prompt $t_m$, VQGAN generates images by optimizing the latent code $z_n$, guided by the semantic relevance score derived from CLIP. Formally,
\begin{align}
    &I_n=\mathcal{E}_{img}(\mathcal{G}(z_n)), n\in[1,N],\\
    &T_m=\mathcal{E}_{txt}(t_m), m\in[1,M],\\
    &\mathcal{L}_{z_n}=\mathcal{D}(I_n, T_m),
\label{invert}
\end{align}where $\mathcal{E}_{img}(\cdot)$ and $\mathcal{E}_{txt}(\cdot)$ denote the image encoder and the text encoder of CLIP, mapping the synthesized image and the text prompt to the image embedding $I_n$ and the text embedding $T_m$, respectively. Similar to \cite{clipvqgan}, $\mathcal{D}(\cdot,\cdot)$ is the squared spherical distance for text-image semantic relevance measurement. For balance training, we synthesize $\frac{N}{M}$ images for each category, where $N$ is the total number of the synthesized images and $M$ is the number of task-specific categories. With the synthesized images, the student can be trained by knowledge distillation, transferring the task-specific knowledge from CLIP:
\begin{align}
    &y_n=[\mathcal{D}(I_n,T_1),...,\mathcal{D}(I_n,T_M)],\\
    &\mathcal{L}_{\theta_S}=\mathcal{L}_{1}(\mathcal{S}(\mathcal{G}(z_n), \theta_S), y_n),
\end{align}where $\mathcal{S}(\cdot, \theta_S)$ is the customized student. 

Conventional DFKD implicitly relies on the premise that the test data involves little domain shift, which is vulnerable in real-world scenarios. In contrast, DFKD-VLFM aims to develop a generalizable student model and delivers it to domain-agnostic downstream tasks. To achieve this, it is crucial to extract comprehensive knowledge of the task-specific label space from CLIP by generating surrogate images that cover a wide range of data distribution. However, since the given information is only in the form of category words, the most intuitive method is to create numerous prompt templates by hand, such as ``[{\tt CLS}] {\tt in the style of} [{\tt STYLE}]''. However, this approach is laborious, empirical, and poses a high risk of model collapse. To address this issue, we propose to find a new word that represents the novel ``style'' of a category and make three attempts at Prompt Diversification.

\subsection{Prompt Diversification}
\subsubsection{\textbf{Mix-Prompt}}
It is intuitive to source the words or phrases about image style from the Internet and build a hand-crafted word dictionary. By stitching different \emph{style} words with the \emph{category} word, we can achieve different sentences of a category. However, the limited size of the dictionary still poses a risk of mode collapse. Hence, we extend it to a broader space via a random convex interpolation. Specifically, we mix-up two different text embedding ($T_{m1}$ and $T_{m2}$) of the same category:

\begin{align}
    &T_{m}=\lambda T_{m1} + (1-\lambda)T_{m2},
\end{align}where $\lambda$ is the coefficient randomly sampled from a \emph{Beta} distribution. $T_{m}$ is the corresponding category text embedding of the novel style.

\subsubsection{\textbf{Random-Prompt}}
\label{Sec:Random-Prompt}
In real-world scenarios, it is essential to synthesize data that covers a wide range of domain distributions. However, it is difficult to capture the entire semantic concept of a category using a limited-scale word dictionary, even when expanded by random convex interpolation. To address this issue, we propose a pseudo word to describe a category with an abstract concept, rather than relying solely on a concrete concept of image style in the Mix-Prompt approach.

Before introducing Random-Prompt, we first review the pre-processing step of the text encoder. Similar to BERT~\cite{Devlin2019BERTPO}, the input string is converted into a set of tokens, each of which is associated with a corresponding token embedding once the text encoder has been trained. This creates a fixed codebook, where each word can be converted into an embedding vector $V$ through index-based lookup in the codebook $C\in \mathbb{R}^{K\times D}$. Here $K$ represents the codebook size and $D$ is the token feature dimension. This step is referred to as \emph{token embedding} in Fig.\ref{fig: framework}. After this, all individual token embeddings of an input string are combined to construct a sentence embedding, also known as text embedding in this paper. Therefore, the text encoder $\mathcal{E}_{txt}$ can be decomposed into a token embedding $\mathcal{E}_{token}$ and a sentence embedding $\mathcal{E}_{sentence}$:

\begin{equation}
    \mathcal{E}_{txt}=\mathcal{E}_{sentence}\circ  \mathcal{E}_{token}.
\end{equation}

As such, we designate a place-holder string, $H_*$, as the pseudo word to represent the novel style. This allows us to generate infinite styles by randomly generating a continuous virtual token embedding $V_*$ for $H_*$. The corresponding token embedding for $H_*$ is denoted as $V_*=\mathcal{E}_{token}(H*)$. Assuming that the token embedding space follows a Multivariate Gaussian distribution, we estimate the statistics of the distribution using token embeddings from the codebook $C$ and generate $V_*$ by randomizing from this distribution:

\begin{align}
    \mu_d&=\frac{1}{K}\sum_{j\in[1,K]} C_{j,d},\\
    \sigma^2_d&=\frac{1}{K}\sum_{j\in[1,K]}C_{j,d}^2 - \mu_d^2.
\end{align}Given the mean $\mu$ and the standard deviation $\sigma$, the token embedding $V_*$ can be randomized by Gaussian sampling and the text embedding $T_{m}$ for image synthesis is achieved:
\begin{align}
    &V_{*,d}=\mu_d + e \cdot \epsilon_d \cdot \sigma_d, \,\,\,\, \epsilon_d\sim \mathcal{N}(0, 1),\\
    &T_{m}=\mathcal{E}_{sentence}([V_m], V_*),
    \label{eq:random-prompt}
\end{align}where $[V_m]$ is the token embedding of the target category name and the frozen template. And $\mathcal{N}(0, 1)$ is the standard Gaussian distribution. $e$ is a scale scalar which can enlarge $\sigma_d$ so as to enrich diversification range.

\subsubsection{\textbf{Contrastive-Prompt}}
Given the extremely vast visual concept space in large-scale vision-language pre-trained models, Mix-Prompt and Random-Prompt may not yield distinguishable instances, even with the infinite virtual styles generated. For example, a dog can appear in various styles, poses, and backgrounds, and even belong to different breeds.  We conjecture that generating each text prompt independently lacks instance-level diversity, making it insufficient to extract generalizable knowledge from such models. To further diversify the data, we propose a learnable prompt optimized by instance-level contrastive learning~\cite{Chen2020ASF}. The core idea is to enlarge the distance between the current prompt and the historical generated ones. Specifically, we adopt a memory bank mechanism to store the historical ones. The details of this process are illustrated in Fig.~\ref{fig: framework}. Initialized by Random-Prompt or Mix-Prompt, we develop a contrastive loss in the text embedding space $T_m$, back-propagating to optimize $V_*$ in the token embedding space:
\begin{equation}
    \mathcal{L}_{V_*}=-log(\frac{exp(\mathcal{D}(T_{m}, T_{m}^+)/\tau)}{\sum_{b\in[1,B]} exp(\mathcal{D}(T_{m}, T_b^-)/\tau)}),
\end{equation}where $\mathcal{D}(\cdot,\cdot)$ is the cosine similarity function and $\tau$ is the temperature parameter. The positive pairs are comprised of the current text prompts and the augmented counterparts $T_m^+$, being optimized to be closer. The negative samples $T_b^-$ are constructed by different text prompts from the current batch and the memory bank, being pulled apart. $B$ is the number of negative samples. In this paper, we adopt a commonly used text augmentation method - randomly shuffling the text letter of the frozen template ``{\tt in the style of}'' to construct the positive samples. Since the word of the target category is kept fixed, the semantic information can be preserved when some letters of the frozen template are shuffled.

\section{Experiments}
In the following sections, we aim to demonstrate the practicality of DFKD-VLFM, where a generalized student is tailored for domain-agnostic downstream tasks without access to real data. We conduct experiments across a wide range of popular domain generalization (DG) datasets using two fundamental image classification settings: 1) Zero-shot learning, to assess whether the knowledge of CLIP can be effectively transferred upon synthetic datasets. 2) Few-shot learning, to evaluate whether the distilled student can serve as a promising pre-trained model for downstream tasks, even surpassing publicly available pre-trained models with the same architecture in an agnostic domain. Further ablation studies and analyses are conducted to validate the efficacy of the proposed approaches.

\subsection{Experiment Settings}
\subsubsection{\textbf{Datasets}}
We use four popular DG datasets to evaluate the out-of-distribution generalization ability of the students on image classification tasks, including \underline{PACS}\cite{pacs} (Art Painting (A), Cartoon (C), Photo (P) and Sketch (S)), \underline{VLCS}\cite{vlcs} (Caltech101\cite{caltech101}, LabelMe\cite{labelme}, SUN\cite{sun}, VOC2007\cite{voc2007}), \underline{ImageCLEF-DA}\cite{imageclef} (Caltech (C), ImageNet (I) and Pascal (P)) and \underline{VisDA}\cite{visda} (Synthetic and real).
\subsubsection{\textbf{Implementation details}}
We adopt ViT-B/32~\cite{Dosovitskiy2021AnII} as the visual encoder and a 12-layer transformer as the textual encoder of CLIP. VQGAN is utilized for surrogate images generation. The pre-trained weights of CLIP\footnote{https://github.com/openai/CLIP} and VQGAN\footnote{https://github.com/CompVis/taming-transformers} are downloaded from the official Github page of the corresponding papers and are kept fixed throughout the training process. Without any specific statements, we use a ResNet18~\cite{resnet} as the default student model for ablation studies.
Specifically, we search from the Internet to develop a style dictionary with the size of 86 to develop Mix-Prompt. As for Contrastive-Prompt, the embedding of the pseudo word is initialized with Random-Prompt and optimized by a contrastive loss. The surrogate images are generated with the resolution $224\times224$. 

We report the mean classification accuracy over five runs with random seeds. Please refer to the Appendix for more details.

\subsection{Zero-shot Classification}
Since CLIP exhibits strong zero-shot performance, it is of interest to determine whether the generalizable knowledge can be transferred to a lightweight student using synthetic data. In this section, we evaluate the performance of the distilled student on several datasets with distinct domain shifts to verify its generalizability. Notably, all the methods described in this section are trained without any real-world data (zero-shot). Given the word ``CLS'' of the target categories, CLIP conducts zero-shot classification with the vanilla prompt ``{\tt A [CLS]}''. The baseline synthesizes the surrogate data using the same vanilla prompt of zero-shot CLIP and performs knowledge distillation upon the synthetic dataset. Table~\ref{tab: PACS}-\ref{tab: ImageCLEF} summarizes the quantitative results, where the student can inherit task-specific knowledge from CLIP to some extent. Introducing Prompt Diversification consistently outperforms the baseline setting by a large margin across all datasets, indicating the importance of diversity in surrogate datasets. Notably, Random-Prompt without prompt engineering can achieve comparable results to Mix-Prompt with elaborate prompt engineering, demonstrating the convenience of Random-Prompt. Contrastive-Prompt achieves the best performance among the proposed three Prompt Diversification methods, especially in the PACS dataset with large domain gaps, where the exhaustive search of diverse styles driven by the contrastive learning paradigm is believed to be responsible for the remarkable improvement. To further evaluate the effectiveness of our method, in Table~\ref{tab: VisDA}, we conduct experiments on VisDA, a larger dataset, where Contrastive-Prompt still achieves satisfactory results.

\begin{table}[!t]
	\caption{Results on PACS. \textbf{Bold} fonts represent the best results. ``Avg.'' denotes the average accuracy. }
	\label{tab: PACS}
\setlength{\tabcolsep}{8pt}
	\centering
	{\resizebox{0.475\textwidth}{!}{
		\begin{tabular}{l|cccc|c}
			\toprule
			Method &P &A &C &S &Avg.\\
			\midrule
			CLIP &99.46 &95.56 &97.44 &84.14 &94.15 \\
			\midrule
			\multicolumn{6}{c}{ResNet18}\\
			\midrule
			Baseline &86.18 &51.21 &48.57 &25.27 &52.81 \\
			Mix-Prompt &90.24 &72.29 &78.17 &50.10 &72.70\\
            Random-Prompt &91.27 &69.26 &77.29 &38.18 &69.00\\
			Contrastive-Prompt &\textbf{93.44} &\textbf{79.74} &\textbf{80.42}  &\textbf{52.78} &\textbf{76.59}\\
			\midrule
			\multicolumn{6}{c}{ResNet50}\\
			\midrule
			Baseline &89.04 &56.08 &54.86 &22.42 &55.60  \\
			Contrastive-Prompt &\textbf{94.69} &\textbf{81.76} &\textbf{81.74} &\textbf{55.04} &\textbf{78.31} \\
			\bottomrule
	\end{tabular}
	}
}
\end{table}

\begin{table}[!t]
	\caption{Results on VLCS. \textbf{Bold} fonts represent the best results. ``Avg.'' denotes the average accuracy.}
	\label{tab: VLCS}
\setlength{\tabcolsep}{8pt}
	\centering
	{\resizebox{0.475\textwidth}{!}{
		\begin{tabular}{l|cccc|c}
			\toprule
			Method &V &L &C &S &Avg.\\
			\midrule
			CLIP &70.65 &56.37 &99.70 &74.27 &75.25 \\
			\midrule
			\multicolumn{6}{c}{ResNet18}\\
			\midrule
			Baseline &46.59 &51.55 &79.01 &54.43 &57.90 \\
			Mix-Prompt &62.46 &61.52 &90.58 &67.96 &70.63 \\
			Random-Prompt &63.46 &\textbf{62.27} &91.00 &69.22 &71.49\\
			Contrastive-Prompt &\textbf{66.55} &61.72 &\textbf{95.52} &\textbf{73.53} &\textbf{74.33}\\
			\midrule
			\multicolumn{6}{c}{ResNet50}\\
			\midrule
			Baseline &54.64 &56.20 &81.72 &62.95 &63.88  \\
			Contrastive-Prompt &\textbf{69.75} &\textbf{62.51} &\textbf{96.02} &\textbf{75.86} &\textbf{76.04} \\
			\bottomrule
	\end{tabular}
	}
}

\end{table}

\begin{table}[!t]
	\caption{Results on ImageCLEF. \textbf{Bold} fonts represent the best results. ``Avg.'' denotes the average accuracy.}
	\label{tab: ImageCLEF}
\setlength{\tabcolsep}{12pt}
	\centering
	{\resizebox{0.475\textwidth}{!}{
		\begin{tabular}{l|ccc|c}
			\toprule
			Method &C &I &P &Avg.\\
			\midrule
			CLIP &94.00 &93.83 &82.33 &90.06 \\
			\midrule
			\multicolumn{5}{c}{ResNet18}\\
			\midrule
			Baseline &79.80 &75.23 &64.27 &73.10 \\
			Mix-Prompt &85.57 &80.60 &65.77 &77.31 \\
            Random-Prompt &86.33 &82.10 &67.23 &78.56\\
			Contrastive-Prompt &\textbf{87.33} &\textbf{84.60} &\textbf{69.33} &\textbf{80.42} \\
			\midrule
			\multicolumn{5}{c}{ResNet50}\\
			\midrule
			Baseline &83.67 &77.70 &68.20 &76.46  \\
			Contrastive-Prompt &\textbf{90.75} &\textbf{87.75} &\textbf{73.00} &\textbf{83.83} \\
			\bottomrule
	\end{tabular}
	}
}

\end{table}

\begin{table*}[]
	\caption{Experimental results on VisDA. ``Avg.'' denotes the average accuracy of twelve categories per domain.}
	\label{tab: VisDA}
    \centering
    {\resizebox{0.95\textwidth}{!}{
    \begin{tabular}{c|l|l|cccccccccccc|c}
         \toprule
			Domain &Model &Method &{\rotatebox{90}{plane}} &{\rotatebox{90}{bcycl}} &{\rotatebox{90}{bus}} &{\rotatebox{90}{car}} &{\rotatebox{90}{horse}} &{\rotatebox{90}{knife}} &{\rotatebox{90}{mcycl}} &{\rotatebox{90}{person}} &{\rotatebox{90}{plant}} &{\rotatebox{90}{sktbrd}} &{\rotatebox{90}{train}} &{\rotatebox{90}{truck}} &Avg.\\
		\midrule
		\multirow{9}*{\rotatebox{90}{Synthetic}}&ViT-B/32&CLIP &92.92 &83.45 &91.69 &69.01 &99.58 &68.57 &99.83 &99.42 &99.66 &84.46 &83.18 &56.77 &85.71 \\
		\cmidrule{2-16}
		&\multirow{4}*{ResNet18}&Baseline &67.11 &82.51 &86.40 &67.73 &98.11 &31.43 &90.66 &65.25 &99.56 &39.41 &79.89 &31.88 &69.99\\
		&&Mix-Prompt &65.05 &85.43 &86.36 &67.65 &98.68 &29.67 &83.97 &57.74 &98.12 &57.84 &68.50 &39.18 &69.85\\
		&&Random-Prompt &54.08 &73.93 &86.69 &74.87 &80.26 &37.40 &89.51 &73.93 &96.33 &36.52 &66.26 &44.47 &68.85  \\
		&&Contrastive-Prompt &66.86 &82.00 &86.50 &72.06 &95.58 &32.74 &91.78 &78.80 &98.80 &26.69 &69.35 &39.80 &70.08 \\
		\cmidrule{2-16}
		&\multirow{2}*{ResNet50}&Baseline &72.11 &80.28 &79.99 &78.42 &100 &37.71 &98.01 &63.12 &99.95 &21.18 &74.18 &40.00 &70.41  \\
		&&Contrastive-Prompt &60.16 &74.45 &86.16 &70.68 &99.28 &41.85 &94.81 &73.55 &99.19 &34.42 &73.26 &35.08 &70.24 \\
		\bottomrule
		\toprule
		\multirow{9}*{\rotatebox{90}{Real}} &ViT-B/32&CLIP &87.64 &85.34 &81.04 &89.16 &90.12 &77.62 &86.76 &67.78 &97.29 &54.47 &91.55 &76.64 &82.12 \\
		\cmidrule{2-16}
		&\multirow{4}*{ResNet18}&Baseline &93.89 &80.45 &77.18 &71.16 &90.24 &31.21 &57.68 &31.04 &91.37 &31.82 &70.01 &60.95 &65.58\\
		&&Mix-Prompt &91.74 &82.15 &69.98 &75.45 &89.94 &39.49 &70.54 &29.06 &95.45 &57.56 &68.98 &60.57 &69.24 \\
		&&Random-Prompt &84.73 &70.80 &60.07 &77.73 &78.44 &47.31 &76.15 &37.81 &93.67 &65.26 &61.03 &57.77 &67.56 \\
		&&Contrastive-Prompt &93.66 &77.80 &69.50 &71.82 &83.81 &43.80 &86.21 &35.21 &92.91 &52.64 &85.45 &52.99 &70.48 \\
		\cmidrule{2-16}
		&\multirow{2}*{ResNet50}&Baseline &86.75 &85.67 &63.73 &77.00 &97.10 &52.37 &64.41 &35.82 &95.88 &25.93 &74.30 &64.94 &68.66\\
		&&Contrastive-Prompt &87.64 &82.13 &71.95 &76.34 &89.40 &60.55 &88.20 &34.08 &95.04 &60.91 &87.22 &61.03 &74.54 \\
		\bottomrule
    \end{tabular}
    }
    }

\end{table*}

\begin{figure}[t]
	\centering
	\includegraphics[width=1\columnwidth]{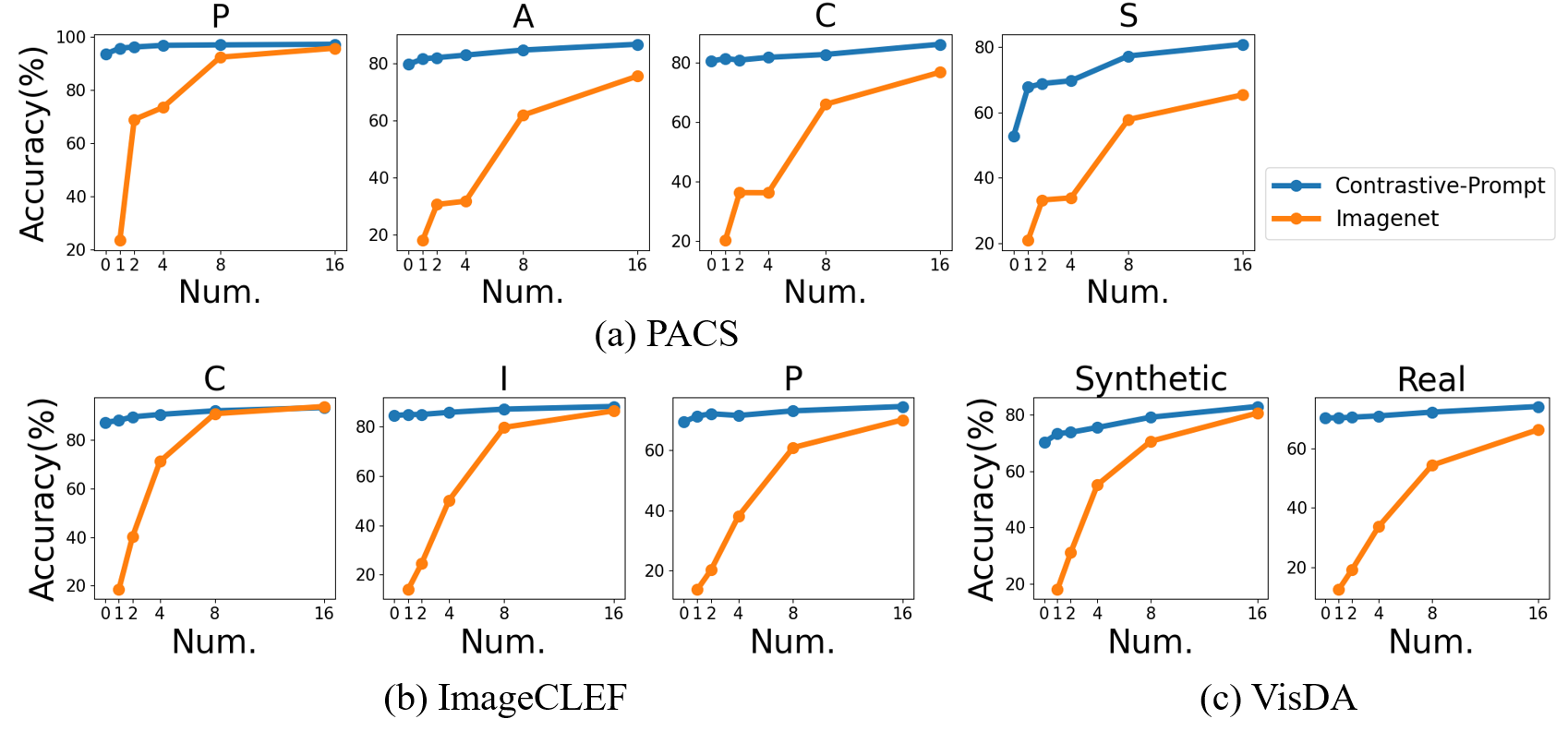} 
	\caption{Results of few-shot fine-tuning on three datasets. ``Num.'' denotes the shots for training. Contrastive-Prompt drives the crafted student into a strong few-shot learner (\textcolor{blue}{blue} lines), transcending the pre-trained model on ImageNet (\textcolor{orange}{orange} lines).}
	\label{fig: fewshots}
\end{figure}

\subsection{Few-shot Fine-tuning}
In this section, we investigate the effectiveness of the distilled model in few-shot setting, which is expected to perform a fast adaptation to the real data distribution as another evidence of the effective generalizable data-free knowledge transfer. Specifically, we conduct few-shot experiments on each domain, following the common few-shot protocols of training on 1, 2, 4, 8, and 16 samples per category, with the remainder of the full dataset serving as the test set. All results represent the average of five runs with random seeds. Performance comparisons are drawn between the Contrastive-Prompt-driven distilled models and ImageNet pre-trained models sharing identical network architectures.

Based on the average performance across domains in all datasets, as illustrated in Fig. \ref{fig: fewshots}, we observe that the grafted student from CLIP is a strong few-shot learner that can quickly adapt to target data with only a few samples and surpass the ImageNet pre-trained models by a substantial margin. These appealing results can support our assumption that large pre-trained vision-language models can serve as effective teachers for customizing specific students for edge devices, inheriting significant generalizability from the vision-language foundation models. Thus, it is feasible to create a more promising pre-trained classifier with arbitrary structure rather than the ImageNet pre-trained models.

\begin{figure*}[!t]
	\centering
	\includegraphics[width=2.1\columnwidth]{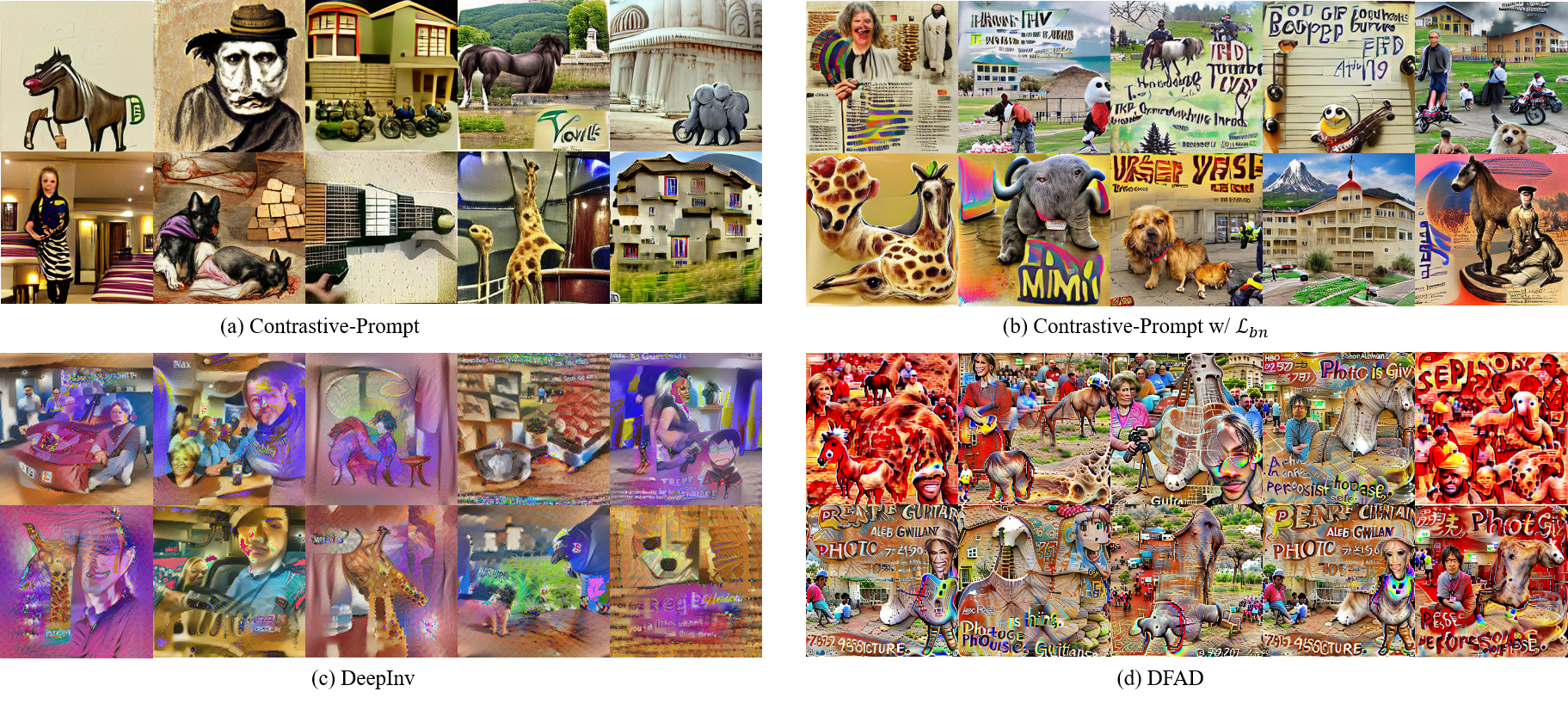} 
	\caption{Qualitative Comparisons between DFKD-VLFM and the conventional DFKD methods. DFKD-VLFM takes text prompt as a bridge for diverse image synthesis, while the conventional DFKD methods aim to directly synthesize diverse images.}
	\label{fig: DFKD}
\end{figure*}

\begin{table}[!t]
	\caption{Comparison with conventional DFKD methods. RN50 is adopted as the visual backbone of CLIP.}
	\label{tab: DFKD}
\setlength\tabcolsep{10pt}
	\centering
	{\resizebox{0.48\textwidth}{!}{
		\begin{tabular}{l|cccc|c}
			\toprule
			Method &P &A &C &S &Avg.\\
			\midrule
            DeepInv\cite{Deepinv} &79.59 &62.91 &54.07 &23.93 &55.13\\
            DFAD\cite{DFAD} &34.49 &31.54 &27.39 &20.31 &28.43\\
            CMI\cite{CMI}&38.68 &39.06 &30.59 &20.82 &32.29 \\
            PRE-DFKD\cite{PRE-DFKD}&27.60&22.36&15.32&15.47&20.19\\
            Contrastive-Prompt w/ $\mathcal{L}_{bn}$ &86.05 &61.60 &68.15 &32.83 &62.16\\
            Contrastive-Prompt &\textbf{87.92} &\textbf{66.86} &\textbf{67.01} &\textbf{45.46} &\textbf{68.81}\\
			\bottomrule
	\end{tabular}}
}

\end{table}

\subsection{Comparison with DFKD Methods}
This section provides an elaborate comparison between the proposed DFKD-VLFM task and the DFKD approaches~\cite{Deepinv,DFAD,CMI,PRE-DFKD}. To this end, we perform experiments on PACS and present both quantitative and qualitative comparisons in Table~\ref{tab: DFKD} and Fig.~\ref{fig: DFKD}, respectively. It is important to note that BN regularization $\mathcal{L}_{bn}$ is a crucial component of DFKD~\cite{Deepinv}, which regularizes the feature statics $(\mu_i(g(z)),\sigma_i(g(z)))$ of synthesized data in the $i_{th}$ BN layer conformed to the original ones $(\hat{\mu}_i,\hat{\sigma}_i^2)$ stored in the pre-trained teacher model. However, it is not applicable to Transformer models. For a fair comparison, we use Resnet-50 as the visual backbone for all experiments in this section.

The results presented in Table~\ref{tab: DFKD} demonstrate that the student model guided by Contrastive-Prompt consistently outperforms the DFKD methods, and the inclusion of $\mathcal{L}_{bn}$ leads to a significant drop in performance. We attribute this poor performance of DFKD to several factors. Firstly, the use of high-fidelity synthetic images significantly facilitates knowledge transfer. As shown in Fig.\ref{fig: DFKD}, the images generated by VQGAN are more realistic and retain the identifying category features, which promotes knowledge transfer from the teacher to the student. On the other hand, although DFAD employs a generator, it is trained from scratch, which increases the risk of mode collapse, as demonstrated in prior research\cite{CMI,largeDFAD}. Secondly, $\mathcal{L}_{bn}$ has an inferior effect on synthetic data. CLIP is pre-trained on a large-scale dataset sourced from the Internet, which has a significant domain shift with respect to the test images that are often complex scenes with multiple objects and textual descriptions. As shown in Fig.\ref{fig: DFKD}(b)-(d), synthetic data generated by $\mathcal{L}_{bn}$ follows the data distribution of web images, which are often corrupted with textual characters, leading to a degradation in performance of the student model. Furthermore, the prior space of the well-pretrained CLIP and VQGAN are fixed during the image generation. Note that $\mathcal{L}_{bn}$ is only related to the priori of CLIP, which enforces the statistics of the synthesized images to conform to the prior statistics stored in the pre-trained CLIP~\cite{Deepinv}. As a result, imposing $\mathcal{L}_{bn}$ narrows the fixed broad prior space and reduces the diversity of the surrogate synthesized dataset, resulting in poorer performance of the distilled student. In contrast, Fig.~\ref{fig: DFKD}(a) depicts more diverse images with fewer textual characters (only one among ten images shown), which implicitly validates the effectiveness of the proposed DFKD-VLFM framework.

\subsection{Ablation Study}
\begin{figure}[t]
	\centering
	\includegraphics[width=0.95\columnwidth]{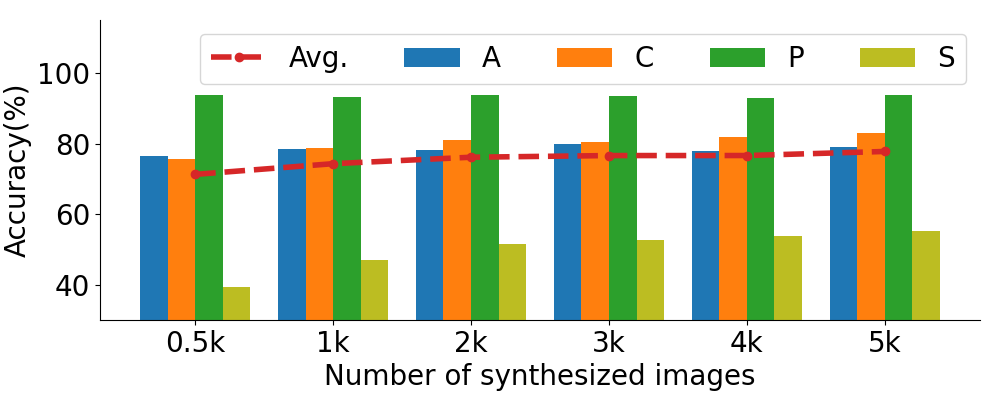} 
	\caption{Ablation study on the scale of synthesized data on PACS. The $x$-axis denotes the image number per class.}
	\label{fig: PACS_numbers}
\end{figure}
\begin{figure}[!t]
	\centering
	\includegraphics[width=0.85\columnwidth]{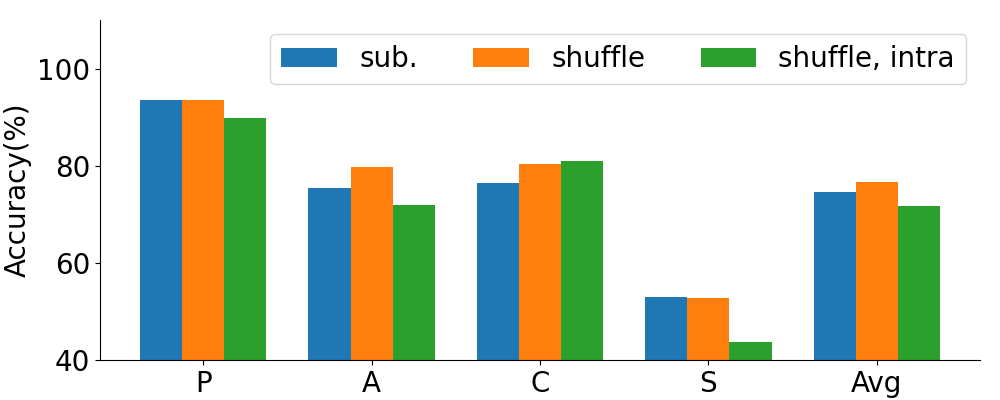}
	\caption{Ablation study on Contrastive-Prompt. ``sub.'' denotes substituting the characters. ``shuffle'' denotes randomly swapping the characters. ``shuffle, intra'' denotes intra-class contrastive learning with character shuffle augmentation.}
	\label{fig: abl contrastive}
\end{figure}
\subsubsection{\textbf{The scale of the synthesized dataset.}} Fig. \ref{fig: PACS_numbers} gives an analysis of the impact of the synthetic dataset scale. The art painting, cartoon, and photo domains enjoy a considerable performance even with limited synthesized images, whereas the sketch domain shows a striking rise in performance as the quantity of synthesized data grows. To balance the trade-off between accuracy and efficiency, we synthesize 3K images per category in this paper.

\subsubsection{\textbf{The effect of the size of the style dictionary in Mix-Prompt}}
To evaluate the effect of the size of the style dictionary, we randomly select several words or phrases from the complete dictionary to form a sub-dictionary. We then construct Mix-Prompt from the sub-dictionary and repeat this process 5 times by randomly sampling sub-dictionaries from the complete style dictionary. Average accuracy is reported in Table~\ref{tab: size of style dictionary}. The average performance across all four domains continuously increases as the size of the dictionary is increased. This inspires us to develop Random-Prompt and Contrastive-Prompt to achieve infinite novel styles.

\subsubsection{\textbf{The effect of the scale scalar $e$ in Random-Prompt}}
As stated in Section~\ref{Sec:Random-Prompt}, the deviation of $\sigma$ is multiplied by a scale scalar $e$ for enriching the diversification. We estimate the effect of $e$ in this section. As shown in Table~\ref{tab: e}, a larger scale scalar $e$ indeed boosts the performance and the performance of the student reaches the peak when $e=10$. Thus, in this paper, $e$ is set to 10 by default.

\begin{table}[!t]
	\caption{Ablation study on the size of the style dictionary. ViT-B/32 is adopted as the visual backbone and ResNet-18 is utilized as the student model. We randomly sample the sub-dictionary to construct Mix-Prompt each run. The average accuracy is reported.}
	\label{tab: size of style dictionary}
\setlength\tabcolsep{12pt}
	\centering
	{\resizebox{0.48\textwidth}{!}{
		\begin{tabular}{l|cccc|c}
			\toprule
			Size &P &A &C &S &Avg.\\
			\midrule
			10 &90.01 &\textbf{73.02} &76.95 &44.84 &71.21\\
			40 &\textbf{91.05} &72.16 &75.39 &47.00 &71.40\\
			86 &90.24 &72.29 &\textbf{78.17} &\textbf{50.10} &\textbf{72.70}\\
			\bottomrule
	\end{tabular}}
}

\end{table}
\begin{table}[!t]
	\caption{Ablation study on the scale scalar $e$.}
	\label{tab: e}
\setlength\tabcolsep{15pt}
	\centering
	{\resizebox{0.48\textwidth}{!}{
		\begin{tabular}{l|cccc|c}
			\toprule
			$e$ &P &A &C &S &Avg.\\
			\midrule
			1 &91.82 &69.09 &76.64 &27.47 &66.26\\
			10 &\textbf{69.26} &\textbf{77.29} &\textbf{91.27} &38.18 &\textbf{69.00}\\
			100 &68.37 &74.77 &90.58 &\textbf{40.24} &68.49\\
			\bottomrule
	\end{tabular}}
}

\end{table}

\subsubsection{\textbf{Text augmentation for Contrastive-Prompt.}} We mainly compare two text augmentation methods, including \emph{character substitution} and \emph{character shuffle} in Fig. \ref{fig: abl contrastive}. While the former involves substituting some characters from the frozen prompt template, the latter entails swapping them. As shown in Fig. \ref{fig: abl contrastive}, the latter shows consistently better performance. Notably, \emph{character shuffle} only has a minimal impact on the text semantics.

\subsubsection{\textbf{Intra-class vs. instance-level Contrastive-Prompt.}}  We provide an ablation study between intra-class and instance-level contrastive prompts. Intra-class contrastive prompts leverage contrastive learning within the same category, while instance-level contrastive prompts employ contrastive learning between different instances. The results in Fig. \ref{fig: abl contrastive} reveal that instance-level contrastive learning (``shuffle'') outperforms the intra-class one (``shuffle, intra''). The utilization of intra-class contrastive learning may pose a potential risk, whereby the style information encoded in the generated text prompts may be linked to the other categories within the downstream task label space. This will lead to image synthesis with category conflict and confuse the knowledge distillation.

\section{Conclusion}
In this paper, we study an important and novel task of creating a generalizable student model from the powerful vision-language foundation models that can be applied to domain-agnostic downstream tasks for zero-shot inference or few-shot fine-tuning. To the best of our knowledge, this is the first attempt to tackle this task. To facilitate the generalization ability transferring from the foundation models to the student model, we propose three novel Prompt Diversification methods,  \emph{i.e.}, Mix-Prompt, Random-Prompt and Contrastive-Prompt, to promote diverse text prompts, which serves as a bridge to facilitate diverse image synthesis. We hope our work will stimulate further interesting research in this area.

\bibliographystyle{ACM-Reference-Format}
\bibliography{sample-base}


\begin{thebibliography}{36}


\ifx \showCODEN    \undefined \def \showCODEN     #1{\unskip}     \fi
\ifx \showDOI      \undefined \def \showDOI       #1{#1}\fi
\ifx \showISBNx    \undefined \def \showISBNx     #1{\unskip}     \fi
\ifx \showISBNxiii \undefined \def \showISBNxiii  #1{\unskip}     \fi
\ifx \showISSN     \undefined \def \showISSN      #1{\unskip}     \fi
\ifx \showLCCN     \undefined \def \showLCCN      #1{\unskip}     \fi
\ifx \shownote     \undefined \def \shownote      #1{#1}          \fi
\ifx \showarticletitle \undefined \def \showarticletitle #1{#1}   \fi
\ifx \showURL      \undefined \def \showURL       {\relax}        \fi
\providecommand\bibfield[2]{#2}
\providecommand\bibinfo[2]{#2}
\providecommand\natexlab[1]{#1}
\providecommand\showeprint[2][]{arXiv:#2}

\bibitem[Binici et~al\mbox{.}(2022)]%
        {PRE-DFKD}
\bibfield{author}{\bibinfo{person}{K. Binici}, \bibinfo{person}{S. Aggarwal}, \bibinfo{person}{N.~T. Pham}, \bibinfo{person}{K. Leman}, {and} \bibinfo{person}{T. Mitra}.} \bibinfo{year}{2022}\natexlab{}.
\newblock \showarticletitle{Robust and Resource-Efficient Data-Free Knowledge Distillation by Generative Pseudo Replay}. In \bibinfo{booktitle}{\emph{Proceedings of the AAAI Conference on Artificial Intelligence}}.
\newblock


\bibitem[Caputo et~al\mbox{.}(2014)]%
        {imageclef}
\bibfield{author}{\bibinfo{person}{Barbara Caputo}, \bibinfo{person}{Henning M{\"u}ller}, \bibinfo{person}{Jesus Martinez-Gomez}, \bibinfo{person}{Mauricio Villegas}, \bibinfo{person}{Burak Acar}, \bibinfo{person}{Novi Patricia}, \bibinfo{person}{Neda Marvasti}, \bibinfo{person}{Suzan {\"U}sk{\"u}darl{\i}}, \bibinfo{person}{Roberto Paredes}, \bibinfo{person}{Miguel Cazorla}, {et~al\mbox{.}}} \bibinfo{year}{2014}\natexlab{}.
\newblock \showarticletitle{ImageCLEF 2014: Overview and analysis of the results}. In \bibinfo{booktitle}{\emph{International Conference of the Cross-Language Evaluation Forum for European Languages}}. Springer, \bibinfo{pages}{192--211}.
\newblock


\bibitem[Chen et~al\mbox{.}(2019a)]%
        {dafl}
\bibfield{author}{\bibinfo{person}{Hanting Chen}, \bibinfo{person}{Yunhe Wang}, \bibinfo{person}{Chang Xu}, \bibinfo{person}{Zhaohui Yang}, \bibinfo{person}{Chuanjian Liu}, \bibinfo{person}{Boxin Shi}, \bibinfo{person}{Chunjing Xu}, \bibinfo{person}{Chao Xu}, {and} \bibinfo{person}{Qi Tian}.} \bibinfo{year}{2019}\natexlab{a}.
\newblock \showarticletitle{Data-free learning of student networks}. In \bibinfo{booktitle}{\emph{Proceedings of the IEEE/CVF International Conference on Computer Vision}}. \bibinfo{pages}{3514--3522}.
\newblock


\bibitem[Chen et~al\mbox{.}(2020a)]%
        {Chen2020ASF}
\bibfield{author}{\bibinfo{person}{Ting Chen}, \bibinfo{person}{Simon Kornblith}, \bibinfo{person}{Mohammad Norouzi}, {and} \bibinfo{person}{Geoffrey~E. Hinton}.} \bibinfo{year}{2020}\natexlab{a}.
\newblock \showarticletitle{A Simple Framework for Contrastive Learning of Visual Representations}.
\newblock \bibinfo{journal}{\emph{ArXiv}}  \bibinfo{volume}{abs/2002.05709} (\bibinfo{year}{2020}).
\newblock


\bibitem[Chen et~al\mbox{.}(2020b)]%
        {chen2020unsupervised}
\bibfield{author}{\bibinfo{person}{Weijie Chen}, \bibinfo{person}{Shiliang Pu}, \bibinfo{person}{Di Xie}, \bibinfo{person}{Shicai Yang}, \bibinfo{person}{Yilu Guo}, {and} \bibinfo{person}{Luojun Lin}.} \bibinfo{year}{2020}\natexlab{b}.
\newblock \showarticletitle{Unsupervised image classification for deep representation learning}. In \bibinfo{booktitle}{\emph{European Conference on Computer Vision}}. Springer, \bibinfo{pages}{430--446}.
\newblock


\bibitem[Chen et~al\mbox{.}(2019b)]%
        {chen2019all}
\bibfield{author}{\bibinfo{person}{Weijie Chen}, \bibinfo{person}{Di Xie}, \bibinfo{person}{Yuan Zhang}, {and} \bibinfo{person}{Shiliang Pu}.} \bibinfo{year}{2019}\natexlab{b}.
\newblock \showarticletitle{All you need is a few shifts: Designing efficient convolutional neural networks for image classification}. In \bibinfo{booktitle}{\emph{Proceedings of the IEEE/CVF Conference on Computer Vision and Pattern Recognition}}. \bibinfo{pages}{7241--7250}.
\newblock


\bibitem[Choi et~al\mbox{.}(2010)]%
        {sun}
\bibfield{author}{\bibinfo{person}{Myung~Jin Choi}, \bibinfo{person}{Joseph~J Lim}, \bibinfo{person}{Antonio Torralba}, {and} \bibinfo{person}{Alan~S Willsky}.} \bibinfo{year}{2010}\natexlab{}.
\newblock \showarticletitle{Exploiting hierarchical context on a large database of object categories}. In \bibinfo{booktitle}{\emph{2010 IEEE computer society conference on computer vision and pattern recognition}}. IEEE, \bibinfo{pages}{129--136}.
\newblock


\bibitem[Crowson et~al\mbox{.}(2022)]%
        {clipvqgan}
\bibfield{author}{\bibinfo{person}{Katherine Crowson}, \bibinfo{person}{Stella Biderman}, \bibinfo{person}{Daniel Kornis}, \bibinfo{person}{Dashiell Stander}, \bibinfo{person}{Eric Hallahan}, \bibinfo{person}{Louis Castricato}, {and} \bibinfo{person}{Edward Raff}.} \bibinfo{year}{2022}\natexlab{}.
\newblock \showarticletitle{Vqgan-clip: Open domain image generation and editing with natural language guidance}.
\newblock \bibinfo{journal}{\emph{arXiv preprint arXiv:2204.08583}}  \bibinfo{volume}{2} (\bibinfo{year}{2022}).
\newblock


\bibitem[Devlin et~al\mbox{.}(2019)]%
        {Devlin2019BERTPO}
\bibfield{author}{\bibinfo{person}{Jacob Devlin}, \bibinfo{person}{Ming-Wei Chang}, \bibinfo{person}{Kenton Lee}, {and} \bibinfo{person}{Kristina Toutanova}.} \bibinfo{year}{2019}\natexlab{}.
\newblock \showarticletitle{BERT: Pre-training of Deep Bidirectional Transformers for Language Understanding}.
\newblock \bibinfo{journal}{\emph{ArXiv}}  \bibinfo{volume}{abs/1810.04805} (\bibinfo{year}{2019}).
\newblock


\bibitem[Dosovitskiy et~al\mbox{.}(2021)]%
        {Dosovitskiy2021AnII}
\bibfield{author}{\bibinfo{person}{Alexey Dosovitskiy}, \bibinfo{person}{Lucas Beyer}, \bibinfo{person}{Alexander Kolesnikov}, \bibinfo{person}{Dirk Weissenborn}, \bibinfo{person}{Xiaohua Zhai}, \bibinfo{person}{Thomas Unterthiner}, \bibinfo{person}{Mostafa Dehghani}, \bibinfo{person}{Matthias Minderer}, \bibinfo{person}{Georg Heigold}, \bibinfo{person}{Sylvain Gelly}, \bibinfo{person}{Jakob Uszkoreit}, {and} \bibinfo{person}{Neil Houlsby}.} \bibinfo{year}{2021}\natexlab{}.
\newblock \showarticletitle{An Image is Worth 16x16 Words: Transformers for Image Recognition at Scale}.
\newblock \bibinfo{journal}{\emph{ArXiv}}  \bibinfo{volume}{abs/2010.11929} (\bibinfo{year}{2021}).
\newblock


\bibitem[Everingham et~al\mbox{.}({[n.\,d.]})]%
        {voc2007}
\bibfield{author}{\bibinfo{person}{M. Everingham}, \bibinfo{person}{L. Van~Gool}, \bibinfo{person}{C.~K.~I. Williams}, \bibinfo{person}{J. Winn}, {and} \bibinfo{person}{A. Zisserman}.} \bibinfo{year}{[n.\,d.]}\natexlab{}.
\newblock \bibinfo{title}{The {PASCAL} {V}isual {O}bject {C}lasses {C}hallenge 2007 {(VOC2007)} {R}esults}.
\newblock \bibinfo{howpublished}{http://www.pascal-network.org/challenges/VOC/voc2007/workshop/index.html}.
\newblock


\bibitem[Fang et~al\mbox{.}(2013)]%
        {vlcs}
\bibfield{author}{\bibinfo{person}{Chen Fang}, \bibinfo{person}{Ye Xu}, {and} \bibinfo{person}{Daniel~N Rockmore}.} \bibinfo{year}{2013}\natexlab{}.
\newblock \showarticletitle{Unbiased metric learning: On the utilization of multiple datasets and web images for softening bias}. In \bibinfo{booktitle}{\emph{Proceedings of the IEEE International Conference on Computer Vision}}. \bibinfo{pages}{1657--1664}.
\newblock


\bibitem[Fang et~al\mbox{.}(2021a)]%
        {FastDFKD}
\bibfield{author}{\bibinfo{person}{G. Fang}, \bibinfo{person}{K. Mo}, \bibinfo{person}{X. Wang}, \bibinfo{person}{J. Song}, \bibinfo{person}{S. Bei}, \bibinfo{person}{H. Zhang}, {and} \bibinfo{person}{M. Song}.} \bibinfo{year}{2021}\natexlab{a}.
\newblock \showarticletitle{Up to 100x Faster Data-free Knowledge Distillation}. In \bibinfo{booktitle}{\emph{Proceedings of the AAAI Conference on Artificial Intelligence}}.
\newblock


\bibitem[Fang et~al\mbox{.}(2019)]%
        {DFAD}
\bibfield{author}{\bibinfo{person}{G. Fang}, \bibinfo{person}{J. Song}, \bibinfo{person}{C. Shen}, \bibinfo{person}{X. Wang}, {and} \bibinfo{person}{M. Song}.} \bibinfo{year}{2019}\natexlab{}.
\newblock \showarticletitle{Data-Free Adversarial Distillation}.
\newblock  (\bibinfo{year}{2019}).
\newblock


\bibitem[Fang et~al\mbox{.}(2021b)]%
        {CMI}
\bibfield{author}{\bibinfo{person}{G. Fang}, \bibinfo{person}{J. Song}, \bibinfo{person}{X. Wang}, \bibinfo{person}{C. Shen}, \bibinfo{person}{X. Wang}, {and} \bibinfo{person}{M. Song}.} \bibinfo{year}{2021}\natexlab{b}.
\newblock \showarticletitle{Contrastive Model Inversion for Data-Free Knowledge Distillation}. In \bibinfo{booktitle}{\emph{IJCAI}}.
\newblock


\bibitem[Fei-Fei et~al\mbox{.}(2006)]%
        {caltech101}
\bibfield{author}{\bibinfo{person}{Li Fei-Fei}, \bibinfo{person}{Robert Fergus}, {and} \bibinfo{person}{Pietro Perona}.} \bibinfo{year}{2006}\natexlab{}.
\newblock \showarticletitle{One-shot learning of object categories}.
\newblock \bibinfo{journal}{\emph{IEEE transactions on pattern analysis and machine intelligence}} \bibinfo{volume}{28}, \bibinfo{number}{4} (\bibinfo{year}{2006}), \bibinfo{pages}{594--611}.
\newblock


\bibitem[Goodfellow et~al\mbox{.}(2014)]%
        {GAN}
\bibfield{author}{\bibinfo{person}{Ian Goodfellow}, \bibinfo{person}{Jean Pouget-Abadie}, \bibinfo{person}{Mehdi Mirza}, \bibinfo{person}{Bing Xu}, \bibinfo{person}{David Warde-Farley}, \bibinfo{person}{Sherjil Ozair}, \bibinfo{person}{Aaron Courville}, {and} \bibinfo{person}{Yoshua Bengio}.} \bibinfo{year}{2014}\natexlab{}.
\newblock \showarticletitle{Generative adversarial nets}.
\newblock \bibinfo{journal}{\emph{Advances in neural information processing systems}}  \bibinfo{volume}{27} (\bibinfo{year}{2014}).
\newblock


\bibitem[He et~al\mbox{.}(2016)]%
        {resnet}
\bibfield{author}{\bibinfo{person}{Kaiming He}, \bibinfo{person}{Xiangyu Zhang}, \bibinfo{person}{Shaoqing Ren}, {and} \bibinfo{person}{Jian Sun}.} \bibinfo{year}{2016}\natexlab{}.
\newblock \showarticletitle{Deep residual learning for image recognition}. In \bibinfo{booktitle}{\emph{Proceedings of the IEEE conference on computer vision and pattern recognition}}. \bibinfo{pages}{770--778}.
\newblock


\bibitem[Hinton et~al\mbox{.}(2015)]%
        {2015KD}
\bibfield{author}{\bibinfo{person}{Geoffrey Hinton}, \bibinfo{person}{Oriol Vinyals}, {and} \bibinfo{person}{Jeff Dean}.} \bibinfo{year}{2015}\natexlab{}.
\newblock \showarticletitle{Distilling the Knowledge in a Neural Network}.
\newblock \bibinfo{journal}{\emph{Computer Science}} \bibinfo{volume}{14}, \bibinfo{number}{7} (\bibinfo{year}{2015}), \bibinfo{pages}{38--39}.
\newblock


\bibitem[Huang et~al\mbox{.}(2022a)]%
        {huang2022transductive}
\bibfield{author}{\bibinfo{person}{Junchu Huang}, \bibinfo{person}{Weijie Chen}, \bibinfo{person}{Shicai Yang}, \bibinfo{person}{Di Xie}, \bibinfo{person}{Shiliang Pu}, {and} \bibinfo{person}{Yueting Zhuang}.} \bibinfo{year}{2022}\natexlab{a}.
\newblock \showarticletitle{Transductive Clip with Class-Conditional Contrastive Learning}. In \bibinfo{booktitle}{\emph{ICASSP 2022-2022 IEEE International Conference on Acoustics, Speech and Signal Processing (ICASSP)}}. IEEE, \bibinfo{pages}{3858--3862}.
\newblock


\bibitem[Huang et~al\mbox{.}(2022b)]%
        {DIST}
\bibfield{author}{\bibinfo{person}{Tao Huang}, \bibinfo{person}{Shan You}, \bibinfo{person}{Fei Wang}, \bibinfo{person}{Chen Qian}, {and} \bibinfo{person}{Chang Xu}.} \bibinfo{year}{2022}\natexlab{b}.
\newblock \showarticletitle{Knowledge distillation from a stronger teacher}.
\newblock \bibinfo{journal}{\emph{arXiv preprint arXiv:2205.10536}} (\bibinfo{year}{2022}).
\newblock


\bibitem[Inoue et~al\mbox{.}(2018)]%
        {Inoue2018CrossDomainWO}
\bibfield{author}{\bibinfo{person}{Naoto Inoue}, \bibinfo{person}{Ryosuke Furuta}, \bibinfo{person}{T. Yamasaki}, {and} \bibinfo{person}{Kiyoharu Aizawa}.} \bibinfo{year}{2018}\natexlab{}.
\newblock \showarticletitle{Cross-Domain Weakly-Supervised Object Detection Through Progressive Domain Adaptation}.
\newblock \bibinfo{journal}{\emph{2018 IEEE/CVF Conference on Computer Vision and Pattern Recognition}} (\bibinfo{year}{2018}), \bibinfo{pages}{5001--5009}.
\newblock


\bibitem[Jia et~al\mbox{.}(2021)]%
        {align}
\bibfield{author}{\bibinfo{person}{Chao Jia}, \bibinfo{person}{Yinfei Yang}, \bibinfo{person}{Ye Xia}, \bibinfo{person}{Yi-Ting Chen}, \bibinfo{person}{Zarana Parekh}, \bibinfo{person}{Hieu Pham}, \bibinfo{person}{Quoc~V. Le}, \bibinfo{person}{Yun-Hsuan Sung}, \bibinfo{person}{Zhen Li}, {and} \bibinfo{person}{Tom Duerig}.} \bibinfo{year}{2021}\natexlab{}.
\newblock \showarticletitle{Scaling Up Visual and Vision-Language Representation Learning With Noisy Text Supervision}. In \bibinfo{booktitle}{\emph{ICML}}.
\newblock


\bibitem[Li et~al\mbox{.}(2017)]%
        {pacs}
\bibfield{author}{\bibinfo{person}{Da Li}, \bibinfo{person}{Yongxin Yang}, \bibinfo{person}{Yi-Zhe Song}, {and} \bibinfo{person}{Timothy~M Hospedales}.} \bibinfo{year}{2017}\natexlab{}.
\newblock \showarticletitle{Deeper, broader and artier domain generalization}. In \bibinfo{booktitle}{\emph{Proceedings of the IEEE international conference on computer vision}}. \bibinfo{pages}{5542--5550}.
\newblock


\bibitem[Luo et~al\mbox{.}(2020)]%
        {largeDFAD}
\bibfield{author}{\bibinfo{person}{Liangchen Luo}, \bibinfo{person}{Mark Sandler}, \bibinfo{person}{Zi Lin}, \bibinfo{person}{Andrey Zhmoginov}, {and} \bibinfo{person}{Andrew Howard}.} \bibinfo{year}{2020}\natexlab{}.
\newblock \showarticletitle{Large-scale generative data-free distillation}.
\newblock \bibinfo{journal}{\emph{arXiv preprint arXiv:2012.05578}} (\bibinfo{year}{2020}).
\newblock


\bibitem[Meng et~al\mbox{.}(2022)]%
        {meng2022attention}
\bibfield{author}{\bibinfo{person}{Rang Meng}, \bibinfo{person}{Xianfeng Li}, \bibinfo{person}{Weijie Chen}, \bibinfo{person}{Shicai Yang}, \bibinfo{person}{Jie Song}, \bibinfo{person}{Xinchao Wang}, \bibinfo{person}{Lei Zhang}, \bibinfo{person}{Mingli Song}, \bibinfo{person}{Di Xie}, {and} \bibinfo{person}{Shiliang Pu}.} \bibinfo{year}{2022}\natexlab{}.
\newblock \showarticletitle{Attention Diversification for Domain Generalization}. In \bibinfo{booktitle}{\emph{European Conference on Computer Vision (ECCV)}}.
\newblock


\bibitem[Mu et~al\mbox{.}(2022)]%
        {Mu2022SLIPSM}
\bibfield{author}{\bibinfo{person}{Norman Mu}, \bibinfo{person}{Alexander Kirillov}, \bibinfo{person}{David~A. Wagner}, {and} \bibinfo{person}{Saining Xie}.} \bibinfo{year}{2022}\natexlab{}.
\newblock \showarticletitle{SLIP: Self-supervision meets Language-Image Pre-training}. In \bibinfo{booktitle}{\emph{ECCV}}.
\newblock


\bibitem[Peng et~al\mbox{.}(2018)]%
        {visda}
\bibfield{author}{\bibinfo{person}{Xingchao Peng}, \bibinfo{person}{Ben Usman}, \bibinfo{person}{Neela Kaushik}, \bibinfo{person}{Dequan Wang}, \bibinfo{person}{Judy Hoffman}, {and} \bibinfo{person}{Kate Saenko}.} \bibinfo{year}{2018}\natexlab{}.
\newblock \showarticletitle{Visda: A synthetic-to-real benchmark for visual domain adaptation}. In \bibinfo{booktitle}{\emph{Proceedings of the IEEE Conference on Computer Vision and Pattern Recognition Workshops}}. \bibinfo{pages}{2021--2026}.
\newblock


\bibitem[Radford et~al\mbox{.}(2021)]%
        {clip}
\bibfield{author}{\bibinfo{person}{Alec Radford}, \bibinfo{person}{Jong~Wook Kim}, \bibinfo{person}{Chris Hallacy}, \bibinfo{person}{Aditya Ramesh}, \bibinfo{person}{Gabriel Goh}, \bibinfo{person}{Sandhini Agarwal}, \bibinfo{person}{Girish Sastry}, \bibinfo{person}{Amanda Askell}, \bibinfo{person}{Pamela Mishkin}, \bibinfo{person}{Jack Clark}, \bibinfo{person}{Gretchen Krueger}, {and} \bibinfo{person}{Ilya Sutskever}.} \bibinfo{year}{2021}\natexlab{}.
\newblock \showarticletitle{Learning Transferable Visual Models From Natural Language Supervision}. In \bibinfo{booktitle}{\emph{ICML}}.
\newblock


\bibitem[Ramesh et~al\mbox{.}(2022)]%
        {DALL-E}
\bibfield{author}{\bibinfo{person}{Aditya Ramesh}, \bibinfo{person}{Prafulla Dhariwal}, \bibinfo{person}{Alex Nichol}, \bibinfo{person}{Casey Chu}, {and} \bibinfo{person}{Mark Chen}.} \bibinfo{year}{2022}\natexlab{}.
\newblock \showarticletitle{Hierarchical text-conditional image generation with clip latents}.
\newblock \bibinfo{journal}{\emph{arXiv preprint arXiv:2204.06125}} (\bibinfo{year}{2022}).
\newblock


\bibitem[Russell et~al\mbox{.}(2008)]%
        {labelme}
\bibfield{author}{\bibinfo{person}{Bryan~C Russell}, \bibinfo{person}{Antonio Torralba}, \bibinfo{person}{Kevin~P Murphy}, {and} \bibinfo{person}{William~T Freeman}.} \bibinfo{year}{2008}\natexlab{}.
\newblock \showarticletitle{LabelMe: a database and web-based tool for image annotation}.
\newblock \bibinfo{journal}{\emph{International journal of computer vision}} \bibinfo{volume}{77}, \bibinfo{number}{1} (\bibinfo{year}{2008}), \bibinfo{pages}{157--173}.
\newblock


\bibitem[Sun et~al\mbox{.}(2022)]%
        {sun2022dynamic}
\bibfield{author}{\bibinfo{person}{Zhishu Sun}, \bibinfo{person}{Zhifeng Shen}, \bibinfo{person}{Luojun Lin}, \bibinfo{person}{Yuanlong Yu}, \bibinfo{person}{Zhifeng Yang}, \bibinfo{person}{Shicai Yang}, {and} \bibinfo{person}{Weijie Chen}.} \bibinfo{year}{2022}\natexlab{}.
\newblock \showarticletitle{Dynamic Domain Generalization}. In \bibinfo{booktitle}{\emph{IJCAI}}.
\newblock


\bibitem[Yao et~al\mbox{.}(2022)]%
        {Yao2022FILIPFI}
\bibfield{author}{\bibinfo{person}{Lewei Yao}, \bibinfo{person}{Runhu Huang}, \bibinfo{person}{Lu Hou}, \bibinfo{person}{Guansong Lu}, \bibinfo{person}{Minzhe Niu}, \bibinfo{person}{Hang Xu}, \bibinfo{person}{Xiaodan Liang}, \bibinfo{person}{Zhenguo Li}, \bibinfo{person}{Xin Jiang}, {and} \bibinfo{person}{Chunjing Xu}.} \bibinfo{year}{2022}\natexlab{}.
\newblock \showarticletitle{FILIP: Fine-grained Interactive Language-Image Pre-Training}.
\newblock \bibinfo{journal}{\emph{ArXiv}}  \bibinfo{volume}{abs/2111.07783} (\bibinfo{year}{2022}).
\newblock


\bibitem[Yin et~al\mbox{.}(2020)]%
        {Deepinv}
\bibfield{author}{\bibinfo{person}{H. Yin}, \bibinfo{person}{P. Molchanov}, \bibinfo{person}{Z. Li}, \bibinfo{person}{J.~M. Alvarez}, \bibinfo{person}{A. Mallya}, \bibinfo{person}{D. Hoiem}, \bibinfo{person}{N.~K. Jha}, {and} \bibinfo{person}{J. Kautz}.} \bibinfo{year}{2020}\natexlab{}.
\newblock \showarticletitle{Dreaming to Distill: Data-free Knowledge Transfer via DeepInversion}. In \bibinfo{booktitle}{\emph{CVPR}}.
\newblock


\bibitem[Yuan et~al\mbox{.}(2021)]%
        {Yuan2021FlorenceAN}
\bibfield{author}{\bibinfo{person}{Lu Yuan}, \bibinfo{person}{Dongdong Chen}, \bibinfo{person}{Yi-Ling Chen}, \bibinfo{person}{Noel C.~F. Codella}, \bibinfo{person}{Xiyang Dai}, \bibinfo{person}{Jianfeng Gao}, \bibinfo{person}{Houdong Hu}, \bibinfo{person}{Xuedong Huang}, \bibinfo{person}{Boxin Li}, \bibinfo{person}{Chunyuan Li}, \bibinfo{person}{Ce Liu}, \bibinfo{person}{Mengchen Liu}, \bibinfo{person}{Zicheng Liu}, \bibinfo{person}{Yumao Lu}, \bibinfo{person}{Yu Shi}, \bibinfo{person}{Lijuan Wang}, \bibinfo{person}{Jianfeng Wang}, \bibinfo{person}{Bin Xiao}, \bibinfo{person}{Zhen Xiao}, \bibinfo{person}{Jianwei Yang}, \bibinfo{person}{Michael Zeng}, \bibinfo{person}{Luowei Zhou}, {and} \bibinfo{person}{Pengchuan Zhang}.} \bibinfo{year}{2021}\natexlab{}.
\newblock \showarticletitle{Florence: A New Foundation Model for Computer Vision}.
\newblock \bibinfo{journal}{\emph{ArXiv}}  \bibinfo{volume}{abs/2111.11432} (\bibinfo{year}{2021}).
\newblock


\bibitem[Zhao et~al\mbox{.}(2022)]%
        {decoupledKD}
\bibfield{author}{\bibinfo{person}{Borui Zhao}, \bibinfo{person}{Quan Cui}, \bibinfo{person}{Renjie Song}, \bibinfo{person}{Yiyu Qiu}, {and} \bibinfo{person}{Jiajun Liang}.} \bibinfo{year}{2022}\natexlab{}.
\newblock \showarticletitle{Decoupled Knowledge Distillation}. In \bibinfo{booktitle}{\emph{Proceedings of the IEEE/CVF Conference on Computer Vision and Pattern Recognition}}. \bibinfo{pages}{11953--11962}.
\newblock


\end{thebibliography}

\balance
\newpage
\appendix
\section{More Details about experiment settings}
\subsection{Datasets}
Four benchmark datasets are utilized for evaluating the generalization ability of the distilled student.
\begin{itemize}
\item \textbf{PACS}\cite{pacs} is a commonly used domain generalization benchmark, consisting of four domains with different types of image, \emph{i.e.}, Art Painting (A), Cartoon (C), Photo (P) and Sketch (S). There are totally 9991 images of 7 categories.

\item \textbf{VLCS}\cite{vlcs} is collected from four photographic datasets, \emph{i.e.}, Caltech101\cite{caltech101}, LabelMe\cite{labelme}, SUN\cite{sun}, VOC2007\cite{voc2007}, consisting of 10,729 images from five categories.

\item \textbf{ImageCLEF-DA}\cite{imageclef} is comprised of twelve categories from three domains, including Caltech (C), ImageNet (I) and Pascal (P), and each class has 50 images.

\item \textbf{VisDA}\cite{visda} is a large-scale dataset designed for domain generalization, which focuses on the domain shift between synthetic and real data. It contains over 280,000 images across twelve categories.

\end{itemize}

\subsection{More Implementation Details}
\paragraph{\textbf{Prompt Diversification.}} We study three types of prompt diversification methods: 1) Mix-Prompt: We search from the Internet to develop a style dictionary with the size of 86, containing the words/phrases to describe image styles. Please refer to Table \ref{tab: style dictionary} for more details about the style dictionary. $\lambda$ is randomly sampled from a Beta distribution $Beta(\alpha,\alpha)$ and $\alpha$ is set 1.0 by default. 2) Random-Prompt: We replace the embedding of the style word with a vector sampled from the estimated Gaussian distribution. 3) Contrastive-Prompt: The embedding of the style word is initialized with Random-Prompt and optimized by a contrastive loss for 100 epochs with Adam optimizer. The learning rate is 0.1 and the batch size for training is 256. The number of negative instances sampled from the memory bank for each batch is set 2048.

\paragraph{\textbf{Surrogate Image Synthesis.}} Adam optimizer with 500 iterations is used to synthesize images with the resolution $224\times224$. The batch size is 8 and the learning rate is 0.15.

\paragraph{\textbf{Knowledge Distillation.}} 
The distillation process is comprised of two steps: warm-up and fine-tuning. We utilize the pre-trained model as the feature extractor and learn the classifier with 150 epochs for warm-up. Then we fine-tune the whole framework with a 0.1 times learning rate of warm-up step for 150 epochs. The student network is trained with an SGD optimizer with a momentum 0.9, and the learning rate of warm-up step is set 0.1. All the experiments are run on a single NVIDIA V100 GPU. We report the mean classification accuracy over five runs with random seeds.

\section{Analysis of Style Dictionary}
Since the large-scale visual-language foundation models are trained on abundant images annotated with different text descriptions, it is insufficient to invert diverse images of a visual concept using a simple text prompt. To diversify the text prompts with minimal human effort, as shown in Table~\ref{tab: style dictionary}, we have devoted to sourcing the words/phases about image style from the Internet and form a style dictionary with the size of 86. As is mentioned in the section of Mix-Prompt in the paper, we randomly sample two styles from the dictionary which are then mixed up in the text embedding space for novel style generation. Moreover, it is worth emphasizing that the style words appeared in the test datasets are not included in the style dictionary so as to mimic the domain-agnostic scenarios. For example, the domain descriptions of PACS ({\tt Photo}, {\tt Art painting}, {\tt Cartoon} and {\tt Sketch}) are not included in the style dictionary. In this way, we can evaluate the generalization ability of the distilled model on the unknown target domains.
\begin{table*}[!t]
\captionof{table}{Style dictionary used in Mix-Prompt.}
\label{tab: style dictionary}
\centering
{\resizebox{1.0\textwidth}{!}{
		\begin{tabular}{cccccccc}
			\toprule
image&graph&picture&figure &diagram&plot&Midcentury&fantasy\\
view&profile&pattern&drawing&chart&Art Nouveau& naive& mysterious\\
Camille Pissarro& Michelangelo Caravaggio&Claude Monet& Edgar Degas&Edvard Munch& Fauvism& Old-School Tattoo& science fiction\\
Futurism& Impressionism&Picasso& Pop Art& Modern art& Surreal Art& Pen and Ink& blurry\\
Sandro Botticelli&oil paints& watercolours&weird bananas&strange colours& binary&pencil illustration&geometric\\
Architectural& Retro&Directoire& Transitional& Empire& International& photorealism& hazy\\
Kerch& digital& daguerreotype& Medical&cult& Ambiguous &primitivism& icon\\
false-color& comic&graphic& film& favicon& macro&psychedelic&mosaic\\
nior&abstract expressionism& vintage& minimalism&long exposure& portrait& screenprint& surrealism\\
graffiti& cubism&infograph& quickdraw& realistic& pixel\_style& woodcut&collage\\
pointillism &wash painting&flat illustration& folk& isometric& charcoal\\
			\bottomrule
	\end{tabular}}
}
\end{table*}
\begin{figure*}[!t]
	\centering
	\includegraphics[width=2.0\columnwidth]{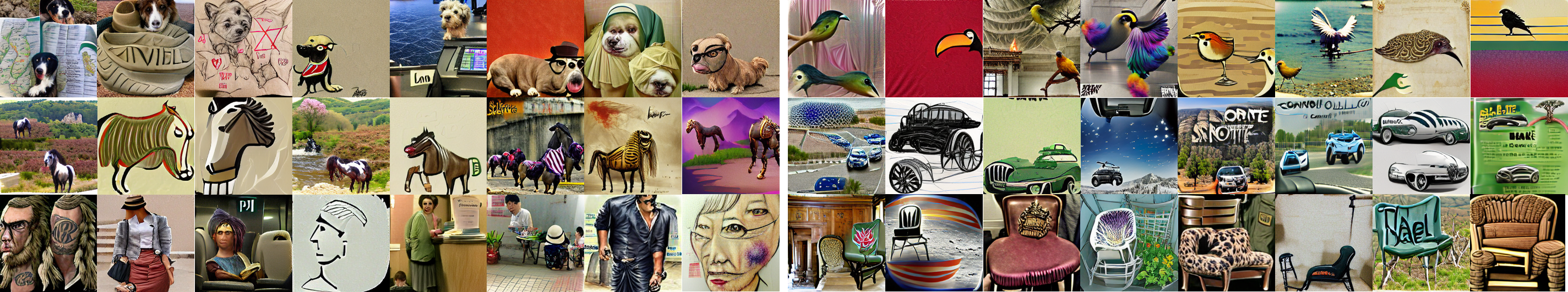} 
	\caption{More visualization obtained by Contrastive-Prompt given a ViT-B/32 as the visual encoder of CLIP. Left: \emph{dog}, \emph{horse}, \emph{person} (from top to bottom). Right: \emph{bird}, \emph{car}, \emph{chair} (from top to bottom).}
	\label{fig: vis2}
\end{figure*}
\begin{figure*}[t]
	\centering
	\includegraphics[width=2\columnwidth]{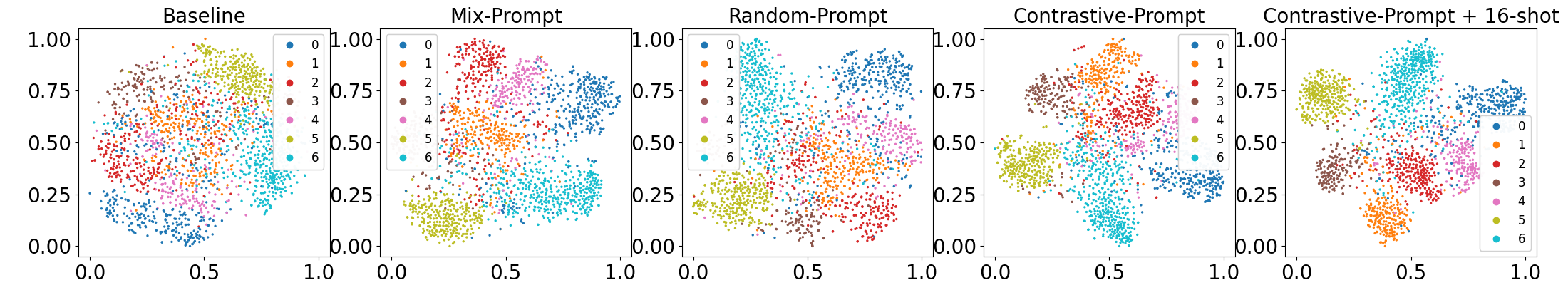} 
	\caption{t-SNE feature visualization on art painting domain of PACS. Different colors denote different categories.}
	\label{fig: tsne}
\end{figure*}
\begin{figure*}[!t]
	\centering
	\includegraphics[width=1.95\columnwidth]{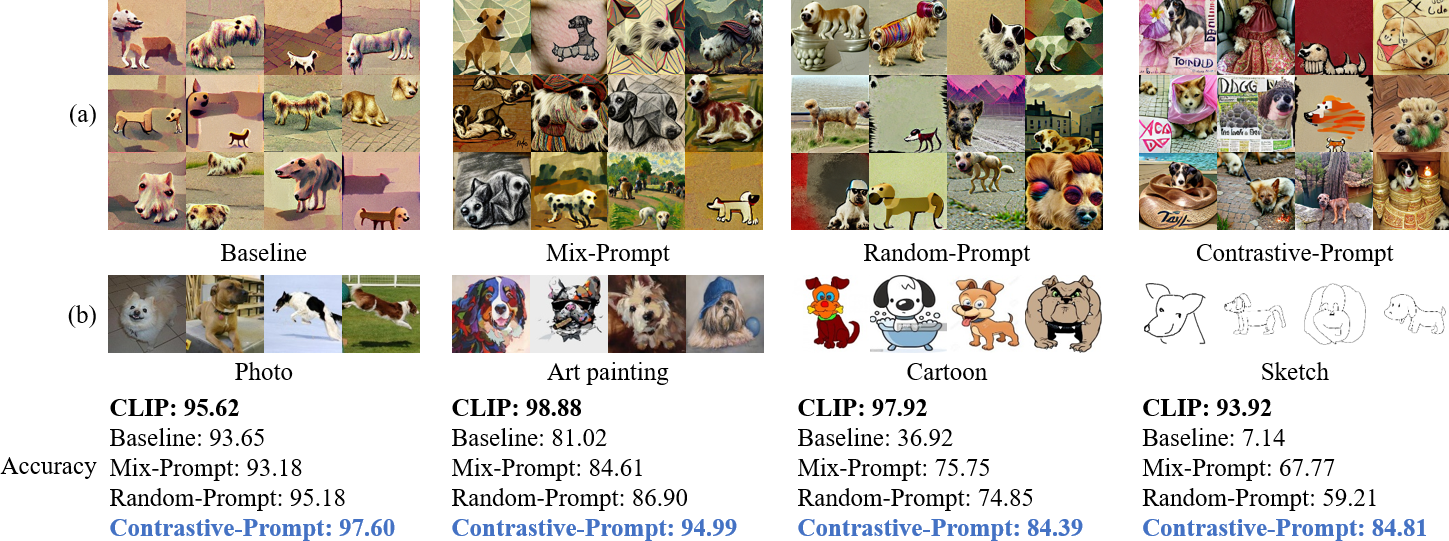} 
	\caption{(a) The images of ``dog'' synthesized by different methods. (b) Real images of ``dog'' from photo, art painting, cartoon and sketch domains of PACS dataset. ``Accuracy'' denotes the performance comparison of the ``dog'' category among different methods on each domain (The results of accuracy from left to right are photo, art painting, cartoon, and sketch).}
	\label{fig: visual of syn}
\end{figure*}
\section{More Visualization of the Synthesized Images}
Fig.~\ref{fig: vis2} gives a more sufficient qualitative visualization of the effectiveness of the proposed Contrastive-Prompt, which enjoys high fidelity and style diversity benefited from the well-pretrained VQGAN and various text prompts. Since a given category can appear in abundant contexts in the agnostic downstream tasks, the diversity of the surrogate synthesized datasets plays a vital importance. Fig.~\ref{fig: vis2} gives a qualitative demonstration of the proposed Contrastive-Prompt in diversifying the text prompts. As shown in the forth-row in Fig.~\ref{fig: vis2}, it is able to render the concept of bird in not only different styles, but also in different breeds and different scenes, while preserving the identifying feature of a bird. We owe this superior property to the contrastive learning performed in the text embedding space, which pulls each instance apart to exhaustively extract target knowledge from the pre-trained CLIP.
\section{More Ablation Studies}
In this section, more ablation studies are carried out on PACS to further evaluate the efficacy of the proposed methods. The setting is exactly the same as the main body of the paper, in which ViT-B/32 is adopted as the visual backbone of CLIP and ResNet18 is adopted as the student model.

\subsection{Feature Visualization.} Fig.\ref{fig: tsne} shows the t-SNE visualization of the features extracted by the distilled model on Art painting domain of PACS. When adapted to the novel domain, we can see the features extracted by the Contrastive-Prompt one are more compact than the others. This gives another evidence that Contrastive-Prompt helps the student learn discriminative features. Given 16 shots for fine-tuning, the features can be fast adapted to the target domain. 

\subsection{Case Study.} We take the visual concept ``dog'' as an example for case study. In Fig.\ref{fig: visual of syn}(a), the proposed Prompt Diversification methods can help synthesize more diverse data than the baseline counterpart, among which Contrastive-Prompt can provide the most diverse images. Fig.\ref{fig: visual of syn}(b) represents different domains of the downstream tasks, which is agnostic during Data-Free Knowledge Distillation. When Generalized to these agnostic domains, Contrastive-Prompt performs the best. Especially in the Sketch domain, the baseline method only achieves 7.14\% accuracy, while Contrastive-Prompt can achieve 84.81\% accuracy, which is superior by a large margin.
\end{document}